\documentclass[letterpaper, 10 pt, journal, twoside]{IEEEtran}

\usepackage{amsmath}
\usepackage{amssymb}
\usepackage{graphicx}
\usepackage{xcolor}
\usepackage[caption=false,font=small]{subfig}

\usepackage{amsthm}
\usepackage{setspace}
\usepackage{hyperref}	
\usepackage{url}
\usepackage{cite}

\graphicspath{{figures/}}
\DeclareGraphicsExtensions{.pdf}

\newcommand{\myws}{4.2cm}
\newcommand{\mywa}{3cm}
\newcommand{\mywb}{2.9cm}
\newcommand{\mywc}{2.7cm}
\newcommand{\myha}{2.45cm}
\newcommand{\myhb}{2.92cm}

\newcommand{\searcher}{s}
\newcommand{\vertex}{v}
\newcommand{\timet}{t}
\newcommand{\deadline}{\tau}
\newcommand{\maxv}{n}
\newcommand{\maxs}{m}
\newcommand{\horizon}{h}
\newcommand{\tset}{T}
\newcommand{\vset}{V}
\newcommand{\sset}{S}
\newcommand{\myset}[2]{#1 = \{1,2,...,#2\}}			% V = {1, 2...n}
\newcommand{\ainA}[2]{#1 \in #2} 					% v \in V
\newcommand{\forallainA}{\forall \ainA}  			% \forall v in V
\newcommand{\halfopen}[2]{\ensuremath{[#1,#2)}}
\newcommand{\mysup}[2]{{#1}^{#2}}
\newcommand{\tos}[1]{\mysup{#1}{s}} 				% x^{s}  
\newcommand{\tot}[2]{{#1}^{t#2}}					% x^{t}
\newcommand{\tost}[2]{{#1}^{s, t#2}} 				% x^{s, t}
\newcommand{\optV}{V^{s, t}}
\newcommand{\bcap}{b_{\rm c}(t)}

\newcommand{\eqc}[1]{Eq. (\ref{#1})}
\newcommand{\eqsc}[2]{Eqs. (\ref{#1})-(\ref{#2})}
\newcommand{\sectc}[1]{Section \ref{#1}}

\newcommand{\target}{object}
\newcommand{\Target}{Object}

\newcommand{\multirobot}{multi-robot}
\newcommand{\Multirobot}{Multi-Robot}
\newcommand{\paper}{letter}

\DeclareMathOperator*{\argmax}{arg\,max}
\newtheorem{theorem}{Theorem}
\newtheorem{corollary}{Corollary}[theorem]

\begin{document}

\title{Mixed-Integer Linear Programming Models for Multi-Robot Non-Adversarial  Search}

% Paper headers
\markboth{IEEE Robotics and Automation Letters. Preprint Version. Accepted June, 2020}
{Asfora \MakeLowercase{\textit{et al.}}: MILP Models for \Multirobot~Non-Adversarial Search}

\author{Beatriz A. Asfora, Jacopo Banfi and Mark Campbell
\thanks{Manuscript received: February, 24, 2020; Revised April, 24, 2020; Accepted June, 23, 2020. This letter was recommended for publication by Editor Nak Young Chong upon evaluation of the reviewers' comments. }
\thanks{All the authors are with the Sibley School of Mechanical and Aerospace Engineering, Cornell University, Ithaca NY 14850, USA. Email: \{ba386, jb2639, mc288\}@cornell.edu}
\thanks{Digital Object Identifier (DOI): 10.1109/LRA.2020.3017473}}

\maketitle

\begin{abstract}
	In this \paper, we consider the \Multirobot~Efficient Search Path Planning (MESPP) problem, where a team of robots is deployed in a graph-represented environment to capture a moving target within a given deadline. We prove this problem to be NP-hard, and present the first set of Mixed-Integer Linear Programming (MILP) models to tackle the MESPP problem. Our models are the first to encompass multiple searchers, arbitrary capture ranges, and false negatives simultaneously. While state-of-the-art algorithms for MESPP are based on simple path enumeration, the adoption of MILP as a planning paradigm allows to leverage the powerful techniques of modern solvers, yielding better computational performance and, as a consequence, longer planning horizons. The models are designed for computing optimal solutions offline, but can be easily adapted for a distributed online approach. Our simulations show that it is possible to achieve 98\% decrease in computational time relative to the previous state-of-the-art. We also show that the distributed approach performs nearly as well as the centralized, within 6\% in the settings studied in this \paper, with the advantage of requiring significant less time -- an important consideration in practical search missions.

\end{abstract}


\section{Introduction}
\label{sec:introduction}

\IEEEPARstart{I}n this letter, we consider the \Multirobot~Efficient Search Path Planning (MESPP) problem introduced by Hollinger et al. in~\cite{hollinger2009}. In this problem, a team of robots is deployed in an environment represented as an undirected graph with the aim of capturing a moving non-adversarial target within a given deadline. In~\cite{hollinger2009}, the authors propose to tackle the MESPP problem with a receding horizon approach that can be implemented either in a centralized or in a distributed fashion. 

In the centralized case, all the possible joint paths are enumerated over a given planning horizon $\horizon$, and the best set of paths is executed until the next planning step. Considering a graph of $\maxv$ vertices and a team of $m$ searchers, this approach has a worst-case complexity of $O(\maxv^{m\horizon})$, i.e. exponential in both the team size and the planning horizon. This approach is dubbed ``explicit coordination''. In the distributed case, instead, at each planning step and for $i=1,\ldots,m$ (according to a lexicographic order), the path of the $i$-th robot is computed by leaving the paths of the remaining robots $j \neq i$ fixed (robots with $j>i$ are initially assumed to remain at their starting position). The optimization of the $i$-th path is again performed by enumerating all feasible paths over a given horizon $\horizon$. This approach is dubbed ``implicit coordination''.

Compared to explicit coordination, implicit coordination scales better w.r.t. the number of robots, with a worst-case complexity of $O(m\maxv^\horizon)$. Implicit coordination also provides an approximation factor $(1 + \kappa)$ along a single planning horizon, where $\kappa$ is the approximation achieved by the solver for the single searcher problem \cite{hollinger2009}, whenever the search objective function can be formulated as a nondecreasing sub-modular set function~\cite{singh2007}. However, the optimization of the single paths still requires the expansion of a search tree with depth $\horizon$. For this reason, practical real-time implementations limit the planning horizon to a few steps ahead (5-6 for a typical indoor environment~\cite{hollinger2009,HollingerThesis}).

{\bf Main contributions.} First, we prove that the MESPP problem is NP-hard even for two-dimensional grid environments with a static target and a single searcher. Second, we present the first set of Mixed-Integer Linear Programming (MILP) models for tackling the MESPP problem, the most general of which is able to encompass multiple searchers, arbitrary capture ranges, and false negatives simultaneously. Our proposed MILP models for the MESPP problem allow path enumeration to be performed in a much more efficient way by leveraging the sophisticated branching and pruning techniques of modern MILP solvers~\cite{bernhard2008}. The models are primarily designed to compute optimal solutions offline, for relatively short missions ($h\leq10$), but they can be used to plan single-robot paths in the same receding-horizon planning scheme introduced in~\cite{hollinger2009}, for longer missions. Our simulation results show that a receding-horizon distributed approach can yield results within 6\% of an optimal offline solution in a matter of seconds. Moreover, the adoption of MILP as a planning paradigm in an online setting can provide 98\% decrease in computational time relative to the state-of-the-art.

This \paper~is structured as follows. \sectc{sec:related_work} frames the MESPP problem within the \multirobot~search literature. \sectc{sec:problemsetting} introduces the MESPP problem and \sectc{sec:nphard} proves its NP-hardness. \sectc{sec:milp} presents the MILP models to compute optimal solutions offline, and \sectc{sec:distributed} shows how the models are adapted for an online distributed implementation. Simulation results are presented in \sectc{sec:simulations}, and \sectc{sec:conclusion} concludes the paper.

\section{Related Work}
\label{sec:related_work}

Target search problems have traditionally been a subject of study in operations research~\cite{stone1976} and game theory~\cite{alpern2006} communities. The past two decades have witnessed an ever-increasing interest in these problems by researchers in mobile robotics, under flavors that make them more suitable to cope with the inherent constraints -- computation, sensing, mobility, communication -- of mobile robots. Chung et al~\cite{chung2011} provide an overview of how target search problems can be tackled from a robotic perspective, introducing a rigorous taxonomy with several classification dimensions. For example, the target can be static~\cite{tisdale2009} or dynamic~\cite{kolling2011}; adversarial ~\cite{kolling2009}, non-adversarial or cooperative~\cite{riccio2016}; in case of known target's motion model, this can be a random walk~\cite{adler2002randomized} or Markovian~\cite{hollinger2009}; the environment can be continuous, unbounded~\cite{alexander} or bounded (typically represented as a polygon~\cite{bhadauria2011}), or discrete and represented by a finite graph~\cite{isler2008}. For what concerns the sensing model, this can be assumed to be perfect, with detection events happening within a given range~\cite{murrieta2007surveillance} or when in line-of-sight with the target~\cite{guibas}, or affected by false negatives~\cite{hollinger2009} and/or false positives~\cite{falsepositives}.

In this paper, we consider the target search problem introduced as MESPP in~\cite{hollinger2009}. This problem version deals with a dynamic, non-adversarial target which moves in a graph-represented environment according to a known Markovian motion model. After its introduction, the MESPP model was used in further studies in data fusion~\cite{hollinger2014} and connectivity problems~\cite{hollinger2012}, most recently in~\cite{banfi2018}. 

The approach presented in~\cite{other_milp} for searching a target on graph-represented environments is also based on a MILP formulation. However, it is restricted to a single searcher and capture events in a single graph vertex. In this letter, we provide a more general formulation able to encompass multiple searchers, arbitrary capture ranges, and false negatives simultaneously.

\section{\Multirobot~Search of a Non-Adversarial \Target}
\label{sec:problemsetting}

This section formalizes the \Multirobot~Efficient Search Path Planning (MESPP) problem. A team of cooperative robots efficiently searches for a non-adversarial target~in a known graph-represented environment within a specified deadline, and the problem goal is to place the searchers where they, as a team, are most likely to intercept the target. The deadline is defined due to a practical reason, i.e., the target~must always be located within a certain time in practical situations. This formulation assumes that the robots have path planning and obstacle avoidance capabilities, allowing them to follow a sequence of waypoints (the graph vertices). Time evolves in discrete steps: each step encompasses the robots' transition between graph vertices, followed by a sensing action at the new vertex. In this \paper, the terms capture, interception, and detection are used interchangeably, as well as the terms object and target.

\subsection{Environment and Searchers' Paths}

Let $G = (V, E)$ be an undirected, connected, and simple graph representing a known environment, with $\myset{\vset}{\maxv}$. The graph can be obtained by discretizing the environment  (for example, a floorplan) by hand, or by means of automated discretization techniques, such as grids~\cite{banfi2018} or constrained Delaunay triangulation~\cite{Delaunay}. We use $\delta(v)$ to denote the neighbors of $\ainA{v}{V}$, while $\delta'(v)$ represents $\delta(v) \cup \{v\}$. Let $d(u,v)$ be the length of the shortest path between any two vertices $u,v \in V$.

Each searcher is represented by $\ainA{\searcher}{\sset}$, where $\myset{\sset}{\maxs}$. Time $\ainA{\timet}{\tset}$ evolves in discrete steps until the deadline $\deadline$, $\myset{\tset}{\deadline}$. Note that in the offline centralized approach, planning horizon $\horizon = \deadline$. A searcher's path $\pi^s$ is defined as an ordered sequence of $\deadline + 1$ vertices $\tos{\pi} = [v_o^s, \mysup{v}{s, 1}, \ldots, \mysup{v}{s, \tau}]$, where $v_o^s$ denotes the starting vertex of $s$. We use $\tos{\mathcal{P}}$ to denote the set of all the possible paths for searcher $s$, and $\mathcal{P}=\prod_{s=1}^{m} \mathcal{P}^s$ to denote the set of all the possible joint paths. Each path must respect the following: at each step, the searcher can either stay still at the current vertex, or move to a neighboring vertex. {Formally, $\forall \{\tost{v}{},\tost{v}{+1}\} \in \pi^s$, $\{\tost{v}{},\tost{v}{+1}\}\in \delta'(v)$.}

\subsection{\Target's Motion and Capture}
\label{sec:targetmotion}

The \target~moves probabilistically in the graph, with motion encoded by a Markov chain specified by the stochastic matrix $\textbf{M} \in \mathbb{R}^{n\times n}$. Specifically, the entry $\textbf{M}_{uv}$ represents the probability that the \target~will move from $u$ to $v$ between time-steps $t$ and $t+1$.

At each step $t$, the \target’s state, resulting from its interactions with searchers executing a set of joint paths $\boldsymbol{\pi} \in \mathcal{P}$, is represented by the belief vector,
\begin{equation}
\textbf{b}^{\boldsymbol{\pi}}(t) = [\bcap,~b_1(t),...,b_{\maxv}(t)] .
\label{eq:beliefvector}
\end{equation}

The first element, $\bcap$, represents the probability that the searchers have located the object by time $t$. The remaining elements $b_1(t),...,b_{\maxv}(t)$ represent the probability that the object is in the corresponding vertices at time $t$, such that $\bcap + \sum_{i=1}^{n}b_{v}(t) = 1$. Note that such probabilities can describe the state of the object in all the possible realizations of the world. In the remaining of this \paper, we will simply denote $\textbf{b}^{\boldsymbol{\pi}}(t)$ by $\textbf{b}(t)$, to reduce the notation burden as the particular set of joint paths will always be clear from the context.

Capture events are described by matrices $\textbf{C}^{s,u} \in [0,1]^{(\maxv+1) \times (\maxv+1)}$, $\forall \searcher \in \sset$, $\ainA{u}{\vset}$. Their effect is to connect the probability of the \target~being at a particular location with its capture state. In other words, the capture matrix $\textbf{C}^{s,u}$ encodes which vertices of the graph fall within the sensing range of searcher $s$, when such is located in vertex $u$.

For the moment, assume the searcher has perfect sensing capabilities. Its capture matrix is constructed as follows. Initialize the capture matrix as an identity matrix $\textbf{C}^{s, u} = \textbf{I}_{n+1}$. Then, for each possible \target~location $v \in V$ that allows a detection when $s$ is placed in $u$, null the $\vertex$-th column of the capture matrix by switching the $1$ at $\textbf{C}^{s,u}_{vv}$ with the $0$ at $\textbf{C}^{s,u}_{v0}$. Note that the first column of $\textbf{C}^{s,u}$ is denoted by index $0$ to avoid confusion with vertex $1$.

Now consider the presence of false negatives in the searcher's sensing actions. False negative rates can vary across the different team members, but we assume they remain constant for each searcher regardless of its position. Let $ \zeta^s \in \halfopen{0}{1}$ be the false negative probability of searcher $s$. When accounting for false negatives, capture matrices are essentially constructed as above: initialize it as an identity matrix $\textbf{I}_{n+1}$, but now replace the $1$ at $\textbf{C}^{s,u}_{vv}$ for $\zeta^s$, and the $0$ at $\textbf{C}^{s,u}_{v0}$ for $1 - \zeta^s$.

The belief update equation links the current belief, the probabilistic \target~motion, and the searchers' paths $\boldsymbol{\pi}=(\pi^1,\ldots,\pi^m)$ with the associated capture events, as follows:

\begin{equation}
\textbf{b}(t+1) = \textbf{b}(t)\left[\begin{array}{cc}
1 & \textbf{0}_{1 \times n} \\ \textbf{0}_{n \times 1} & \textbf{M}\end{array} \right] \prod_{s=1}^{m} {\textbf{C}}^{s, \pi^{s, t+1}}.
\label{eq:beliefupdate}
\end{equation}

\subsection{Optimization Problem}

The MESPP problem seeks to optimize the following objective, subject to \eqc{eq:beliefupdate}:

\begin{equation}
\boldsymbol{\pi}^* = \argmax_{\boldsymbol{\pi} \in \mathcal{P}} \sum_{t=0}^{\deadline} \gamma^t \bcap,
\label{eq:optmax}
\end{equation}
\noindent where $\gamma \in (0,1]$ is a discount factor.

\section{NP-Hardness}
\label{sec:nphard}

In this section we show the hardness of the MESPP problem by proving that its decision version, which we dub MESPP-D, is NP-hard even when the graph is a two-dimensional grid, the target is stationary, and there is a single searcher. We reduce from the Hamiltonian-Path Between Two Points (2HP), proven in~\cite{itai1982} to be NP-complete on grid graphs\footnote{Itai et al.~\cite{itai1982} define grid graphs as finite, vertex induced subgraphs of an infinite graph where (a) the vertex set consists of all points of the plane with integer coordinates and (b) two vertices are connected by an edge if and only if the Euclidean distance between them is equal to 1.}.

\vspace{2mm}
\noindent \textbf{MESPP-D}\\
\noindent INSTANCE: MESPP instance with reward function $F_\tau = \sum_{t=0}^{\deadline} \gamma^t \bcap$, and bound value $B \geq 0$.\footnote{As customarily done, we assume that all the numbers used in the instance are rational~\cite{madani2003undecidability}.} \\
\noindent QUESTION: Is there a search plan such that $F_\tau \geq B$?

\vspace{2mm}
\noindent \textbf{HAMILTONIAN PATH BETWEEN 2 POINTS (2HP)}\\
\noindent INSTANCE: Graph $G = (V, E)$, vertices $v_A$ and $v_B$.\\
\noindent QUESTION: Does $G$ contain a Hamiltonian path beginning with $v_A$ and ending with $v_B$?

\begin{theorem} MESPP-D is NP-hard even when the following conditions hold simultaneously:
\item 1. $G$ is a grid graph, 
\item 2. the target is stationary, and
\item 3. there is only one searcher with perfect sensing capabilities.
\end{theorem}

\begin{proof} We reduce 2HP to MESPP-D in polynomial time as follows. The MESPP-D \textit{grid graph} is the original 2HP {grid graph} $G = (V,E)$ with vertices labeled in a lexicographic order, restricting only $v_1 = v_A$, $v_n = v_B$. We place  at $v_1$ a \textit{single searcher} with perfect sensing capabilities (no false negatives) and able to capture the target only from its current vertex. The \textit{target is stationary}, $\textbf{M} = \textbf{I}_{n}$. The initial belief is set to $b_{\rm c} = b_1 = 0$,  $b_n = \dfrac{1}{(n-1)^2}$ and $b_v = \dfrac{1}{n-1} + \dfrac{1}{(n-1)^2}, ~2 \leq v \leq n-1$. We define the deadline $\tau = n-1$ and bound value $B = \sum_{v=1}^{n} \gamma^v \sum_{u=1}^{v} {b_u}$. The discount factor value does not influence the proof, and is arbitrarily chosen as $\gamma = 0.99$. 
	
Stating $G$ has a Hamiltonian path between vertices $v_A$ and $v_B$ means that it is possible to start at $v_A$, pass through all vertices exactly once and reach $v_B$ within $n-1$ steps.  Note that in the transformation to MESPP-D, $v_B$ maps to the lowest non-zero probability vertex $v_n$, as $b_n < b_v, \text{for}~ v=2,...,n-1$. Thus the search plan with the maximum capture probability by deadline (Eq. \ref{eq:optmax}) requires the searcher to visit each vertex exactly once and lastly the vertex with lowest probability of reward ($v_n$). Following this path yields a reward of exactly $F_\tau = B$, and MESPP-D is therefore a yes-instance.

Conversely, if MESPP-D is a yes-instance, i.e. $F_\tau \geq B$, the searcher must have started at a particular vertex $v_1$ and reached the lowest probability vertex $v_n$ at the last time step.  Otherwise the searcher would have collected a lower reward by the deadline, $F_\tau < B$, due to visiting a vertex more than once (zero reward) or visiting the vertex with the lowest probability early on (cumulative effect of collecting a smaller reward at $t<\tau$). This implies that if $F_\tau \geq B$, 2HP must be a yes-instance.\end{proof}

\begin{corollary}For a stationary target and perfect sensing, MESPP-D is NP-complete.
\end{corollary}

\begin{proof}
	For a graph with $n$ vertices, a searcher needs $n^2$ steps to visit all vertices in an arbitrary order. In the case of a static target, the searcher will have collected all the possible reward by $t=n^2$ even if $\tau > n^2$. Thus the solution depends only on $n$ and not on $\tau$. The solution is of polynomial size and verifiable in polynomial time, placing the problem in NP.
\end{proof}

\section{Mixed-Integer Linear Programming Models}
\label{sec:milp}

This section presents three MILP models for solving the MESPP problem defined in \sectc{sec:problemsetting}. Legal searchers' paths and object's motion are modeled in Sections~\ref{subsec:paths} and \ref{subsec:target_motion}, respectively. These first sets of variables and constraints are common across all models. We then introduce the constraints for different types of capture events. In particular, capture events limited to the same graph vertex without false negatives are presented in Section~\ref{subsec:binary}; capture events with arbitrary capture range are shown in \sectc{subsec:range}; finally, capture events with arbitrary range and false negatives are introduced in \sectc{subsec:false_negatives}.

\subsection{Legal Paths for Searchers}
\label{subsec:paths}

We use $\optV = \{(v,t) \in \vset \times \tset~|~d(v_o^s,v) \leqslant t, ~ \ainA{s}{S}\}$ to denote, for each searcher $s$, the set of all the possible $(v,t)$ states that are compatible with its starting position $v_o^s$ (i.e. the searcher can actually be in $v$ at time $t$). With a slight abuse of notation, we also define $\optV (t) = \{v \in V ~|~(v,t)\in \optV \}$ and $\optV (v) = \{t \in \tset ~|~(v,t)\in \optV\}$. Let us also introduce a dummy goal vertex $\textit{v}_g$, which can be thought of as connecting to all vertices of $G$ in a fictitious manner. 

Consider two sets of binary variables: $\tost{x}{}_v$ denotes the presence of searcher $\searcher$ in vertex $\vertex$ at time $\timet$, and $\tost{y}{}_{uv}$ conveys the fact that searcher $\searcher$ will move from $u$ to $v$ between steps $\timet$ and $\timet+1$. The following constraints enforce the legality of the paths for the searchers: 

\begin{equation}
\mysup{x}{s, 0}_{v^s_o} = \hspace{-3mm} \sum_{j\in \delta'(v^s_o)} \hspace{-1mm} \mysup{y}{s, 0}_{v^s_o j} = \hspace{-3mm} \sum_{\substack{j\in \optV(\deadline)}} y^{s,\deadline}_{jv_g} = 1, ~ \forall s \in S,
\label{eq:startvertex}	
\end{equation}
\begin{equation}
\begin{aligned}
\tost{x}{}_v \hspace{-1mm} = \hspace{-4mm} \sum_{\substack{j\in \delta'(v) \\ \cap \optV(t-1)}} \hspace{-3mm} \tost{y}{-1}_{jv} = \hspace{-2mm} \sum_{\substack{i\in \delta'(v)}} \hspace{-2mm} \tost{y}{}_{vi}, ~ \forall (v,t < \tau) \in \optV,
\end{aligned}
\label{eq:pathconsistency}
\end{equation}
\begin{equation}
\begin{aligned}
x_v^{s, \deadline} = \hspace{-3mm} \sum_{\substack{j\in \delta'(v) \\ \cap \optV(\tau-1)}} \hspace{-3mm} y^{s, \tau -1}_{jv} =  y^{s, \tau}_{v v_g}, ~ \forall v \in V^{s,t}(\tau)~.
\end{aligned}
\label{eq:pathconsistency_2}
\end{equation}

Equations (\ref{eq:startvertex}) set the searchers' starts and goal vertices, while \eqsc{eq:pathconsistency}{eq:pathconsistency_2} ensure path consistency. The variables are formally defined as 
\begin{equation}
\tost{x}{}_v \in \{0,1\}, ~ \forall s \in S,  (v,t) \in V^{s, t},
\label{eq:xv}
\end{equation}
\begin{equation}
\tost{y}{}_{uv} \in \{0,1\}, \forall s \in S, {(u,t < \tau) \in V^{s,t}(t), v \in \delta'(u)},
\label{eq:yv}
\end{equation}
\begin{equation}
{y}^{s, \deadline}_{uv_g} \in \{0,1\}, ~ \forall s \in S, u \in \optV(\tau).
\label{eq:endlegalpaths}
\end{equation}

\subsection{Object's Motion}
\label{subsec:target_motion}

We introduce two sets of continuous variables: $\beta^{t}_i$, representing the entries of the belief vector at time $t$, and $\alpha^{t}_v$, representing the result of the application of the \target's motion model. The constraints below respectively set the initial belief and evolve the \target's location based on the previous belief:
\begin{equation}
\beta_{i}^{0} = b_i(0),~ \forall i \in V \cup \{\rm c\},
\label{eq:initbelief}
\end{equation}
\begin{equation}
\tot{\alpha}{}_{v} = \sum_{u \in V} \textbf{M}_{uv}\tot{\beta}{-1}_{u}, ~ \forallainA{v}{\vset}, \ainA{\timet}{\tset}.
\label{eq:targetmotion}
\end{equation}

The variables are formally defined as
\begin{equation}
\tot{\beta}{}_i \in [0,1],~\forall i \in V \cup \{c\},~t \in \{0\} \cup \tset,
\label{eq:variablecapture}
\end{equation}
\begin{equation}
\tot{\alpha}{}_v \in [0,1],~ \forall v \in V,~t \in \tset .
\label{eq:alpha}
\end{equation}

\subsection{Capture Events in Same Vertex, Binary Detection }
\label{subsec:binary}

Define same-vertex capture with binary detection (0 or 1) as the searcher being in vertex $\vertex$ at time $\timet$, and able to determine with certainty if the \target~is also in $v$. This entails the following property: if no searcher is at vertex $\vertex$, no new information is available about that vertex, and the belief in $\vertex$ is simply the probability that the \target~might have moved there between $\timet-1$ and $\timet$, denoted by \eqc{eq:targetmotion}. On the other hand, if there is at least one searcher at $\vertex$ and no \target~was detected, one can infer the \target~is not in $v$. 

Define then, for each time-step, a binary variable $\tot{\psi}{}_v$ that equals one if and only if there is at least one searcher located in $\vertex$ at time $t$. The belief vector entries can be expressed as
\begin{equation}
\tot{\beta}{}_v = \tot{\alpha}{}_v \left(1 - \tot{\psi}{}_v  \right), ~ \forallainA{t}{\tset}, \ainA{v}{\vset},
\label{eq:nonlinearbelief}
\end{equation}
\noindent which translates to  $\tot{\beta}{}_v = \tot{\alpha}{}_v$ if $\tot{\psi}{}_v = 0$, or  $\tot{\beta}{}_v = 0$ if $\tot{\psi}{}_v = 1$.

The above constraint is nonlinear and can not be applied directly in a MILP model, but it can be formulated in a linearized manner~\cite{glover1975}. The following constraints substitute Eq.~(\ref{eq:nonlinearbelief}) for the belief,
\begin{equation}
\tot{\beta}{}_v  \leqslant 1 - \tot{\psi}{}_v, ~ \forall v \in V,~\ainA{t}{\tset},
\label{eq:linearbelief}
\end{equation}
\begin{equation}
\tot{\beta}{}_v \leqslant \tot{\alpha}{}_v, ~ \forall v \in V,~t \in \tset ,
\end{equation}
\begin{equation}
\tot{\beta}{}_v \geqslant \tot{\alpha}{}_v-\tot{\psi}{}_v,~ \forall v \in V,~t \in \tset,
\end{equation}
\begin{equation}
\tot{\psi}{}_v \in \{0,1\},~ \forall v \in V,~t \in \tset .
\label{eq:psi}
\end{equation}
The relationship between the capture variable $\tot{\psi}{}_v$ and the searchers' positions variables $\tost{x}{}_v$ is expressed as 
\begin{equation}
\sum_{\substack{s \in S~s.t.\\ v \in V^{s, t}{(t)}}} \tost{x}{}_v \leqslant m \tot{\psi}{}_v, ~\forall v \in V,~t \in \tset,
\label{eq:captureconstraint0}
\end{equation}
\begin{equation}
\tot{\psi}{}_v  \leqslant \sum_{\substack{s \in S~s.t.\\ v \in V^{s, t}{(t)}}} \tost{x}{}_v,~ \forall v \in V,~ t \in \tset,
\label{eq:captureconstraint_0}
\end{equation} 
which means that, if all searchers are in vertex $v$ at time $\timet$, $\tot{\psi}{}_v = 1$, and the sum of the searchers positions $\tost{x}{}_v ~ \forallainA{\searcher}{\sset}$ equals the number of searchers in the team, $\maxs\tot{\psi}{}_v $. If however no searcher is in $v$, $\tot{\psi}{}_v = 0$ and so is the sum of $\tost{x}{}_v ~\forall \ainA{\searcher}{\sset}$.
Finally, the probability of the \target~being intercepted within step $\timet$~is the remaining probability after the belief update on all vertices,
\begin{equation}
\beta_c^t = 1-\sum_{v \in V} \tot{\beta}{}_v,~ \forall t \in \tset.
\label{eq:beliefcapture}
\end{equation}

\subsection{Capture Events Within Given Range, Binary Detection}
\label{subsec:range}

Generalizing the capture event, let us say that a searcher positioned in vertex $u$ is able to detect, with certainty, the presence of the \target~in vertex $v$ located within some arbitrary capture range. As before, this assumption entails that the team does not gain additional knowledge about the \target's true position unless the latter can be intercepted. This rationale is again expressed by Eq. (\ref{eq:nonlinearbelief}) and linearized in Eqs. (\ref{eq:linearbelief})-(\ref{eq:psi}). Equations \eqref{eq:captureconstraint0}-\eqref{eq:captureconstraint_0} must, however, be replaced by the following:
\begin{equation}
\sum_{s \in S} \sum_{\substack{u \in V^{s, t}{(t)} \\ s.t.~\textbf{C}^{s,u}_{v0}=1}} \tost{x}{}_u \leqslant m \tot{\psi}{}_v \quad \forall v \in V,~t \in \tset,
\label{eq:captureconstraint1}
\end{equation}
\begin{equation}
\tot{\psi}{}_v  \leqslant \sum_{s \in S} \sum_{\substack{u \in V^{s, t}{(t)} \\ s.t.~\textbf{C}^{s,u}_{v0}=1 }} \tost{x}{}_u \quad \forall v \in V,~ t \in \tset.
\label{eq:captureconstraint2}
\end{equation}

Now, when $v$ is within range, $\tot{\psi}{}_v = 1$ and the sum of $\tost{x}{}_u$ such that $\textbf{C}^{s, u}_{v0} = 1 ~ \forall \searcher \in \sset$ is at most the number of searchers, or $\maxs \tot{\psi}{}_v$. The opposite logic also applies. The probability of the \target~being intercepted within step $\timet$ is enforced by Eq. \eqref{eq:beliefcapture}.

\subsection{Capture Events with False Negatives}
\label{subsec:false_negatives}

In order to account for the false negatives in detection (see Sec.~\ref{sec:problemsetting}), one must modify \eqc{eq:nonlinearbelief}. First, consider a team of only one searcher. If the capture range can reach a particular vertex, the probability the \target~might be there is no longer zero, but rather $\tot{\beta}{}_v = \zeta\tot{\alpha}{}_{v}$, to account for the chance that the \target~is actually in that vertex, but has not been detected. If the searcher can not reach vertex $\vertex$, no new information is available, and the belief is the probability the target has moved there, as before. 

Considering the capture variable $\tot{\psi}{}_v$ as previously presented, the belief update equation for one searcher becomes
\begin{equation}
\tot{\beta}{}_{v} = \left(1-\zeta \right)\tot{\alpha}{}_{v}\left(1-\tot{\psi}{}_{v}\right) + \zeta \tot{\alpha}{}_{v}.
\label{eq:betafalse1}
\end{equation}

For multiple searchers, however, the detection uncertainty must decrease as more robots are in locations where they can potentially detect the \target. This is expressed by updating the belief in an iterative manner, one searcher at a time. To this aim, we define capture and belief variables $\tost{\psi}{}_v$ and $\tost{\beta}{}_v$ for each searcher, and impose the following constraints:
\begin{equation}
\begin{aligned}
\tost{\beta}{}_{v} = \left(1-\zeta^s \right)\beta_{v}^{s-1, t} \left(1 - \tost{\psi}{}_v\right) + \zeta^{s} \beta_{v}^{s-1, t}, \\ \forall s \in S, t \in T, v \in V,
\label{eq:nonlinearbetam}
\end{aligned}
\end{equation} 
\noindent where 
\begin{equation}
\beta^{0, t}_v =  \tot{\alpha}{}_v,~\forallainA{t}{\tset}, v \in \vset.
\label{eq:beta0fn}
\end{equation}
The variables are formally defined as
\begin{equation}
\tost{\beta}{}_v \in [0,1],~\forallainA{t}{\tset}, v \in \vset, s \in \sset,
\end{equation}
\begin{equation}
\tost{\psi}{}_v \in \{0, 1\},~ \forallainA{t}{\tset}, v \in \vset, s \in \sset. \label{eq:psi_s_def}
\end{equation}
To linearize \eqc{eq:nonlinearbetam}, we define the auxiliary variable,
\begin{equation}
\tost{\delta}{}_v = \beta_{v}^{s-1, t} \left(1 - \tost{\psi}{}_v\right),
\end{equation}
\begin{equation}
\tost{\delta}{}_v \in [0, 1] \quad \forallainA{t}{\tset}, v \in \vset, s \in \sset, \label{eq:delta_s_def}
\end{equation}
\noindent and use the same technique as before to yield linear constraints:
\begin{equation}
\tost{\delta}{}_{v} \leqslant 1 - \tost{\psi}{}_{v},~\forall s \in S, t \in T, v \in V,
\end{equation}
\begin{equation}
\tost{\delta}{}_{v} \leqslant \beta_{v}^{s-1, t},~\forall s \in S, t \in T, v \in V,
\end{equation}
\begin{equation}
\tost{\delta}{}_{v} \geqslant \beta_{v}^{s-1, t} - \tost{\psi}{}_{v},~ \forall s \in S, t \in T, v \in V.
\end{equation}

Equation~\eqref{eq:nonlinearbetam} can therefore be rewritten as
\begin{equation}
\tost{\beta}{}_{v} = \left(1-\zeta^s \right)\tost{\delta}{}_v + \zeta^s \beta_{v}^{s-1, t}, ~\forallainA{t}{\tset}, v \in \vset, s \in \sset.
\label{eq:betafalsem}
\end{equation}

The capture events must now be expressed separately for each searcher:
\begin{equation}
\sum_{\substack{u \in V^{s, t}{(t)} \\ s.t.~\textbf{C}^{s,u}_{v0}>0}} \tost{x}{}_u \leqslant \tost{\psi}{}_v \quad \forallainA{t}{\tset}, v \in \vset, s \in \sset,
\end{equation}
\begin{equation}
\tost{\psi}{}_v  \leqslant \sum_{\substack{u \in V^{s, t}{(t)} \\ s.t.~\textbf{C}^{s,u}_{v0}>0}} \tost{x}{}_u \quad \forallainA{t}{\tset}, v \in \vset, s \in \sset.
\label{eq:capturefn}
\end{equation}

Equation~\eqref{eq:beliefcapture} is again used to express the probability that the \target~has been captured by time $t$, noting that

\begin{equation}
\tot{\beta}{}_v = \beta^{m, t}_v,~\forallainA{t}{\tset}, v \in \vset.
\label{eq:beliefm}
\end{equation}

\subsection{Complete MILP Models}
\label{sec:milpfull}

For same-vertex capture, no false negatives:
\begin{align}
\text{(SV-MILP)} & \quad \text{max} \quad \sum_{t \in \tset} \gamma^t \bcap \quad \text{s.t.} \nonumber \\
\text{Eqs.}~ & \text{\eqref{eq:startvertex}-\eqref{eq:alpha}, \eqref{eq:linearbelief}-\eqref{eq:beliefcapture}} \nonumber. 
\end{align} 
For arbitrary capture range, no false negatives:
\begin{align}
\text{(MV-MILP)} & \quad \text{max} \quad \sum_{t \in \tset} \gamma^t \bcap \quad \text{s.t.} \nonumber \\
\text{Eqs.}~\text{\eqref{eq:startvertex}-\eqref{eq:alpha},}& \text{ \eqref{eq:linearbelief}-\eqref{eq:psi}, \eqref{eq:beliefcapture}-\eqref{eq:captureconstraint2}} \nonumber. 
\end{align}
Finally, the most general model, encompassing arbitrary capture ranges and false negatives:
\begin{align}
\text{(FN-MV-MILP)} & \quad \text{max} \quad \sum_{t \in \tset} \gamma^t \bcap \quad \text{s.t.} \nonumber \\
\text{Eqs.}~\text{\eqref{eq:startvertex}-\eqref{eq:alpha},}& \text{ \eqref{eq:beta0fn}-\eqref{eq:psi_s_def}, \eqref{eq:delta_s_def}-\eqref{eq:beliefm}}. \nonumber
\end{align}

\section{Distributed Online Implementation}
\label{sec:distributed}

The MILP models presented on \sectc{sec:milp} are primarily designed to compute optimal or near-optimal solutions offline. However, for large values of mission deadline $\tau$ and team size $m$, a centralized approach does not scale well computationally.

An implicit coordination approach is inherently more scalable than explicit coordination. Recall the algorithm proposed in \cite{hollinger2009}: at each planning step and for $i=1,\ldots,m$ (according to a lexicographic order), the path of the $i$-th robot is computed by leaving the paths of its teammates $j \neq i$ fixed; robots with $j>i$ are initially assumed to remain at their starting position, i.e., $\pi^{j, t} = v^s_o, ~\forall t=1,\ldots,h$. In~\cite{hollinger2009}, the paths of the single robots are optimized by enumeration. Exploring the same search space by leveraging modern solver techniques ensures better scalability in general. To this aim, we can easily adapt our a models for this task. We adopt the implicit coordination algorithm of \cite{hollinger2009}, using our MILP-based approach to perform a more efficient iterative optimization of single paths.

We can solve a sequence of $m$ models, one for each searcher, while assigning a deterministic value to the variables associated with the paths of the teammates. When planning the path for searcher $i$, a deterministic value is assigned for $\forall j \neq i \in \sset$:

\begin{equation}
{x}^{j, t}_v = \left\{ \begin{array}{l}
1, ~\text{iff } v = \pi^{j, t}  \\
0, ~\text{otherwise}.
\end{array} \right.
\end{equation}

A distributed approach decreases the number of variables to be optimized on the MILP model and thus the complexity of the problem, which generally yields a smaller solution time than a centralized planning scheme for $m > 1$. Although communication constraints are not addressed here, these could also be incorporated into the proposed model by leveraging recent work which proposes MILP-based approaches to connected multi-robot path planning~\cite{banfi2018,banficarpin}.
\section{Simulations}
\label{sec:simulations}

\subsection{General Setup}
\label{sec:setupsim}

We use GUROBI \cite{gurobi2019} to solve the MILP models\footnote{Code is open source and available at \url{https://github.com/basfora/milp\_mespp.git}} on a machine equipped with Intel-Core i9-9900K and 32 GB RAM. The maximum number of used threads is set to eight and the presolve level is kept as default (automatic). The solver timeout is set to 30 min for the offline (centralized) approach, and 10 sec for the distributed. Except when stated otherwise, default values are used for the remaining parameters. 

\begin{figure}[ht]
	\centering
	\includegraphics[width=4cm, height=2cm]{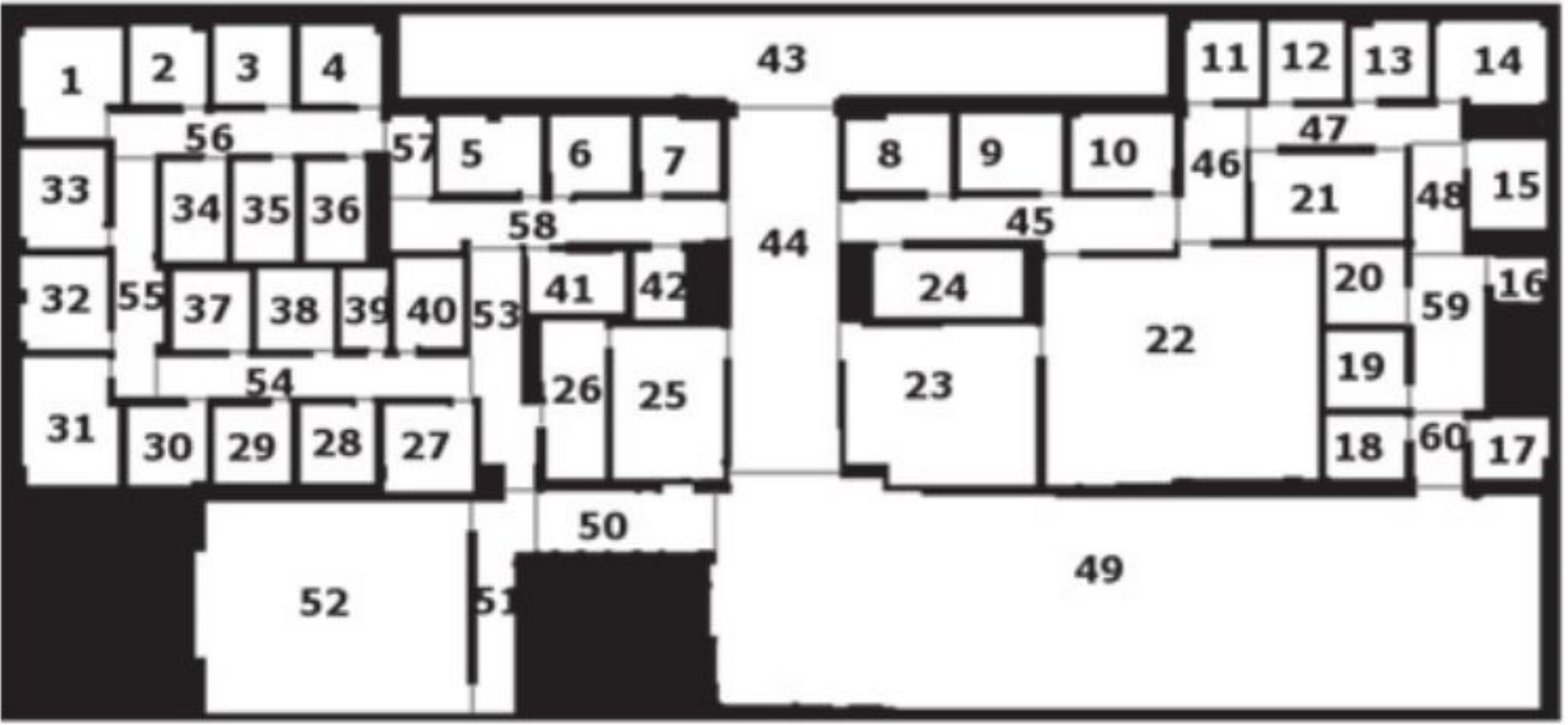}
	\includegraphics[width=4cm, height=2cm]{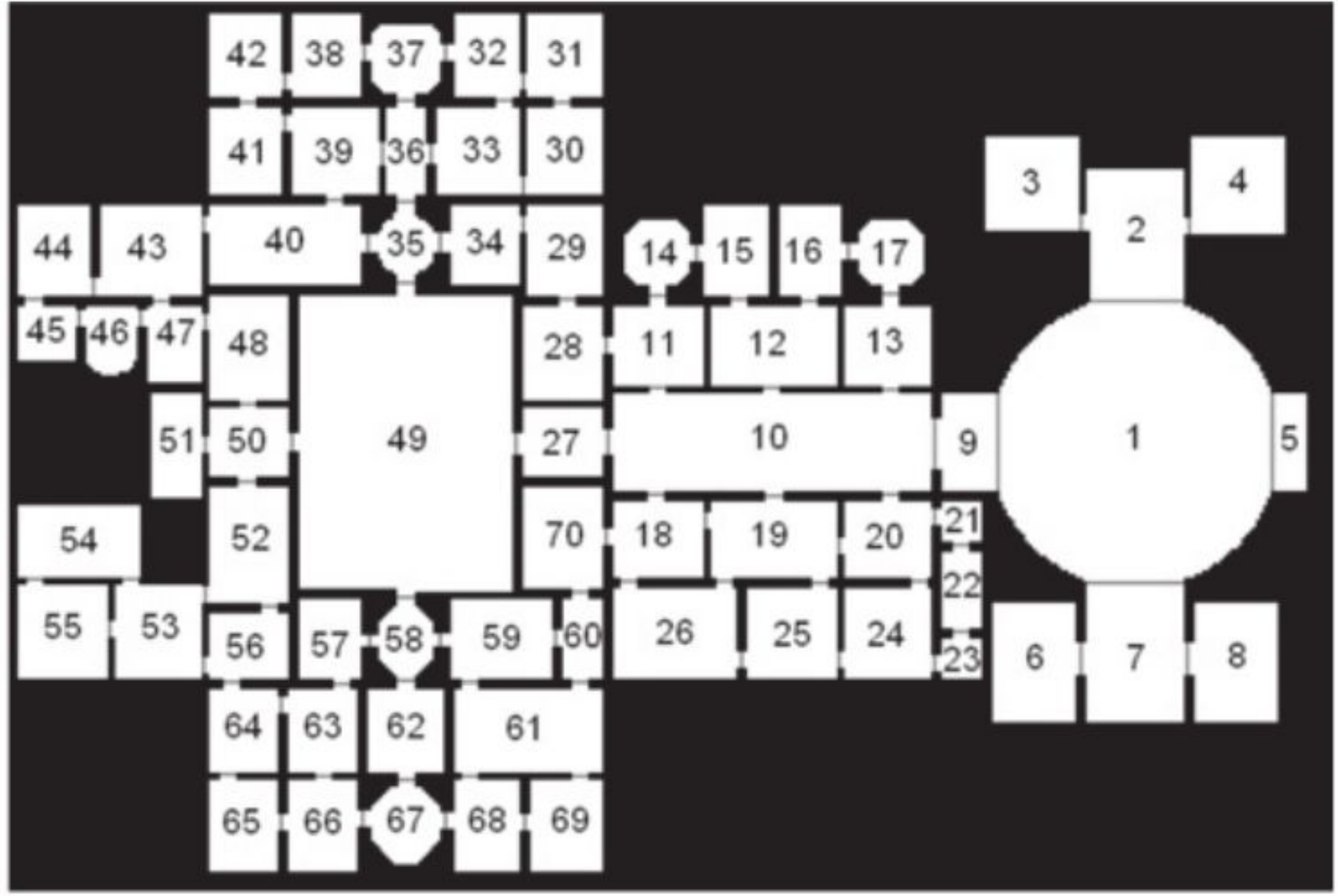}
	\caption{Environments from \cite{hollinger2009} used in evaluations, each room is associated with the corresponding vertex number. Left: OFFICE; Right: MUSEUM.}
	\label{fig:mygraphs}
\end{figure}

We consider three graph environments: OFFICE and MUSEUM, both used in \cite{hollinger2009} and shown in Fig.~\ref{fig:mygraphs}; and a 10x10 4-connected GRID graph.  For OFFICE and MUSEUM, we assume perfect sensing capability and same-vertex capture. For the GRID environment, we assume that the robots' sensing range spans the current vertex plus its 1-hop neighbors, and consider two settings: without (GRID-NOFN) and with (GRID-FN) false negatives, with $\zeta^s=0.3,~\forall s \in S$ for the latter. We use SV-MILP on MUSEUM and OFFICE, MV-MILP on GRID-NOFN, and FN-MV-MILP on GRID-FN.

We perform \textbf{five experiment sets}, each consisting of 100 instances per environment, except for set $4$ in which we double this quantity. The initial configurations (searchers and \target~positions) are randomly chosen. In all instances, the initial capture belief is zero, i.e., the searchers are not able to detect the \target~at the start of the mission. For the initial \target's location belief, we assume an uniform probability between an assorted number of vertices, chosen randomly. For experiment sets $1-3, 5$, there are five possible initial vertices; for experiment set $4$, we vary the number of vertices randomly from two to fifteen; particularly for GRID-FN in set $5$, we consider four possible vertices, drawn from each of the $3\times3$ corner regions of the grid graph, while the initial position of the searchers is drawn from the central portion of the grid.

\subsection{Results}

% ----------------------------------------------------
% SET 1: central, m =1, h={6,..,20}
% ----------------------------------------------------
\subsubsection{Scalability of the MILP models for centralized approach w.r.t. planning horizon length} 
\begin{figure}[h!]
	\centering
	\includegraphics[width=\mywa, height=\myha]{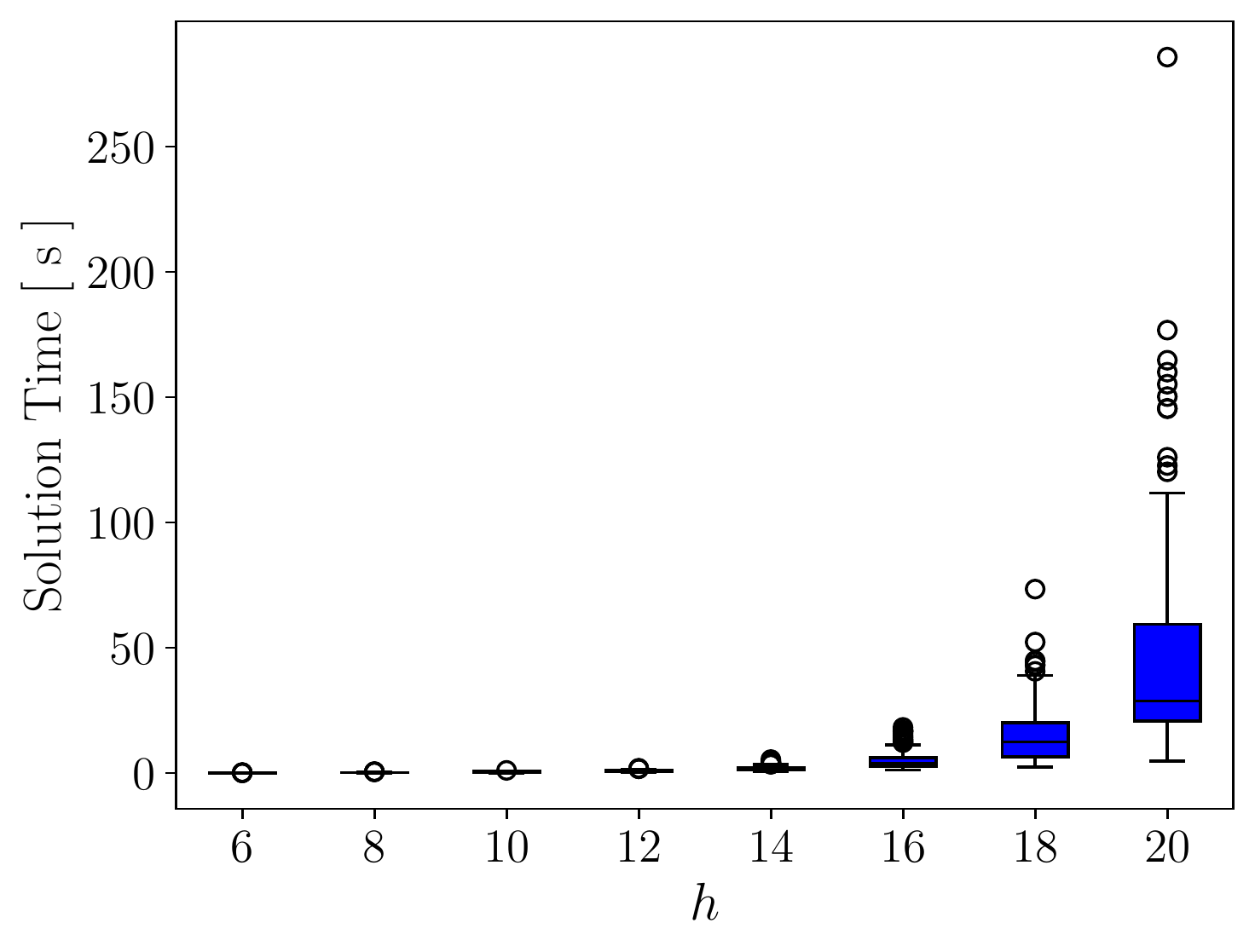}{}
	\includegraphics[width=\mywb, height=\myha]{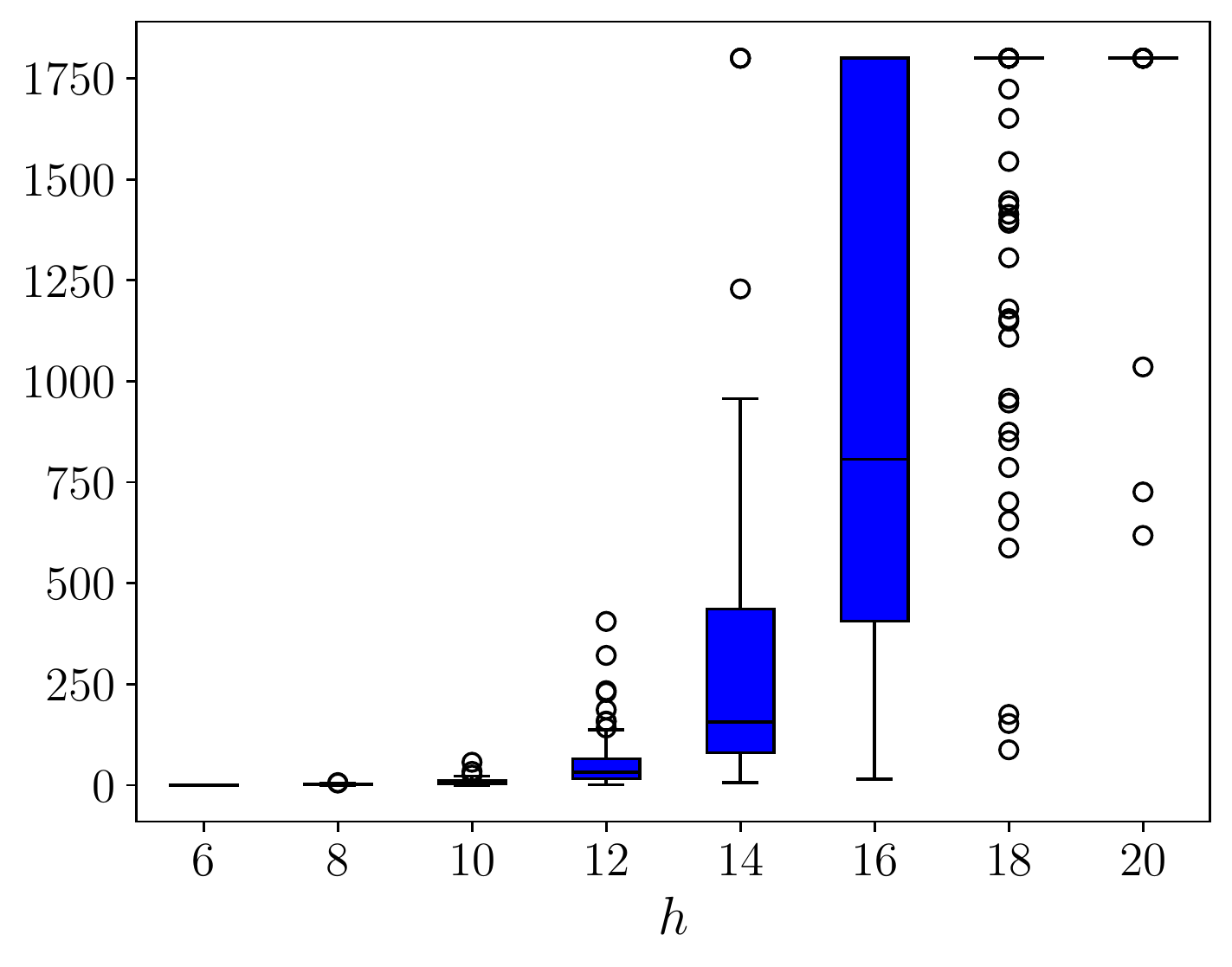}{}
	\includegraphics[width=\mywc, height=\myha]{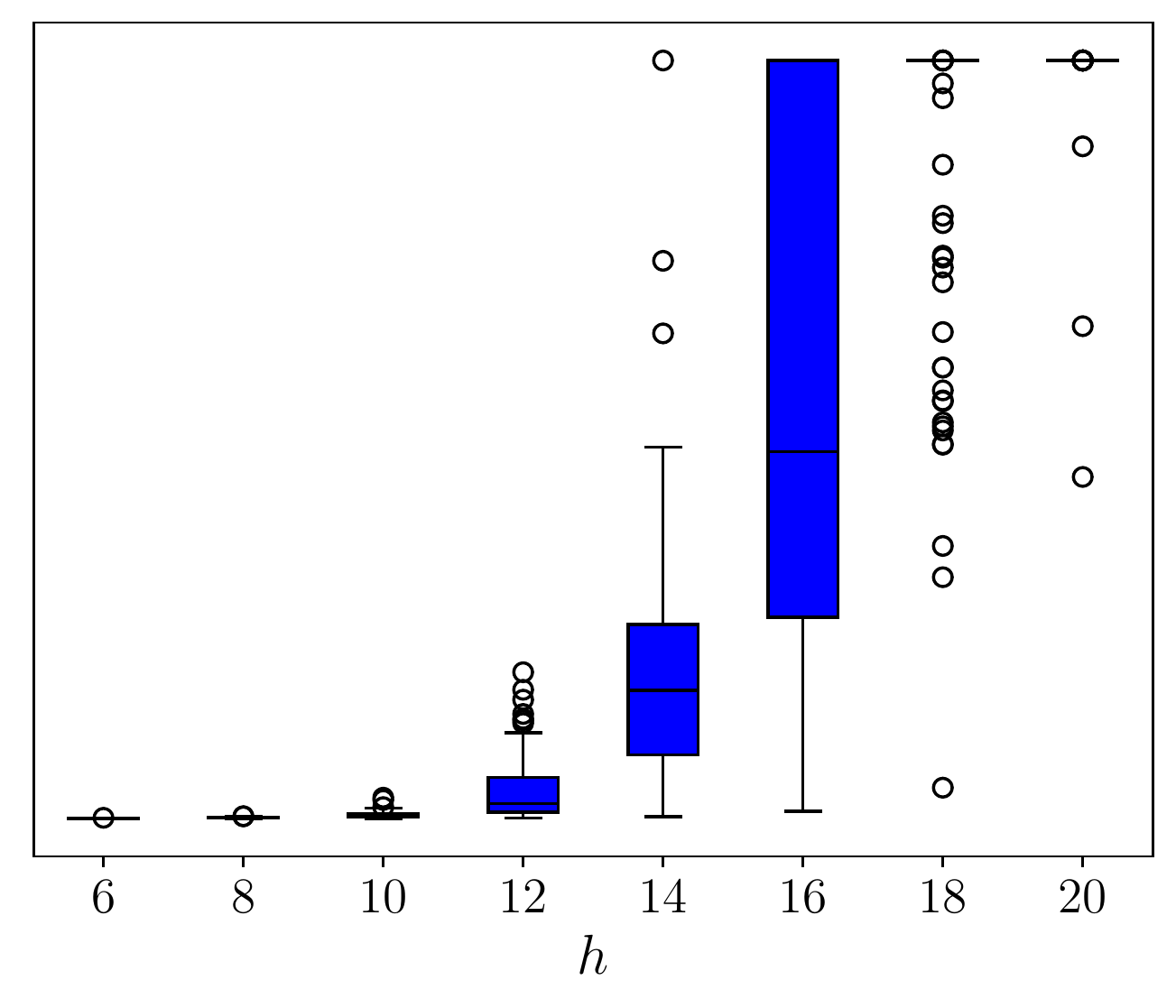}{}
	\caption{Solution times with MILP centralized approach, $m=1$ and varying $h$. Left: OFFICE. Center: GRID-NOFN. Right: GRID-FN.}
	\label{fig:vary_h}
\end{figure}

Fig.~\ref{fig:vary_h} shows the solution times for one searcher ($m=1$) and varying horizons $h$, for OFFICE, GRID-NOFN and GRID-FN.  

The complexity of the model, and therefore the time required to solve it, increases with the complexity of the graph, the robots' sensing range, and the presence of false negatives. OFFICE instances are simpler than GRIDs due to the smaller average graph degree, plus capture events limited to the same vertex. As a result, all OFFICE instances are solved to optimality within a couple of minutes (see Fig.~\ref{fig:vary_h}, left). On the other hand, GRID instances with $h > 16$ tend to hit the solver time limit before an optimal solution can be found. In these sub-optimal cases, the median MIP gap values (not shown in plot) for GRID-FN and $h = 18, 20$ were respectively $11\%$ and $23\%$. The presence of false negatives in the centralized approach does not have a significant impact for $m=1$ (Fig. \ref{fig:vary_h}~center, right). As defined in \eqc{eq:nonlinearbetam}, additional intermediate variables are necessary for each searcher and time-step, causing the increase in complexity to become relevant for multiple searchers. This is confirmed by experiment set 2 (Figs. \ref{fig:m_gaps}-\ref{fig:m_times}).

% ----------------------------------------------------
% SET 2 central, m ={1,..5}, h=10
% ----------------------------------------------------
\subsubsection{Performance of the centralized MILP approach for different team sizes} 
\begin{figure}[ht]
	\centering
	\includegraphics[width=\mywa, height=\myha]{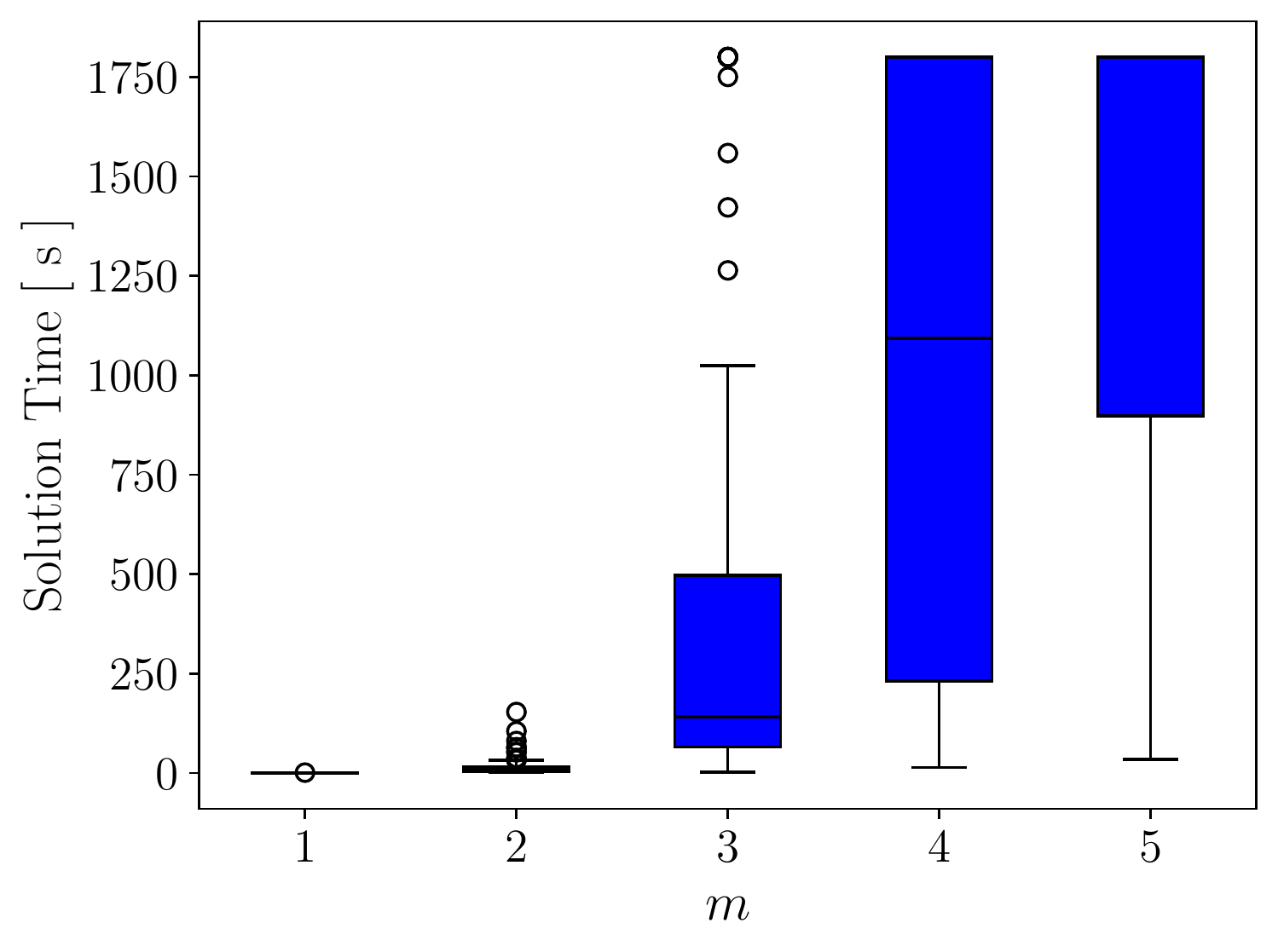}{}
	\includegraphics[width=\mywb, height=\myha]{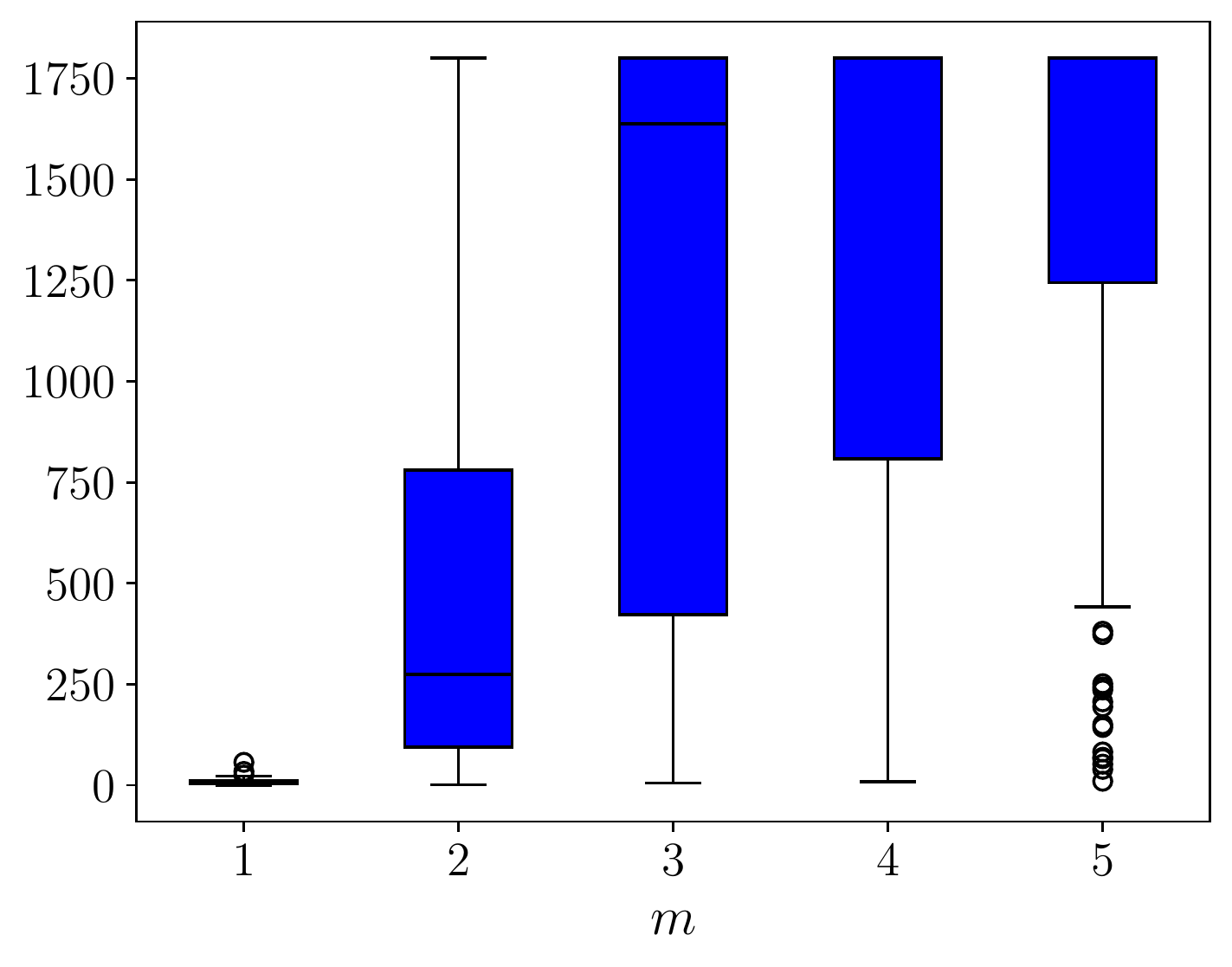}{}
	\includegraphics[width=\mywc, height=\myha]{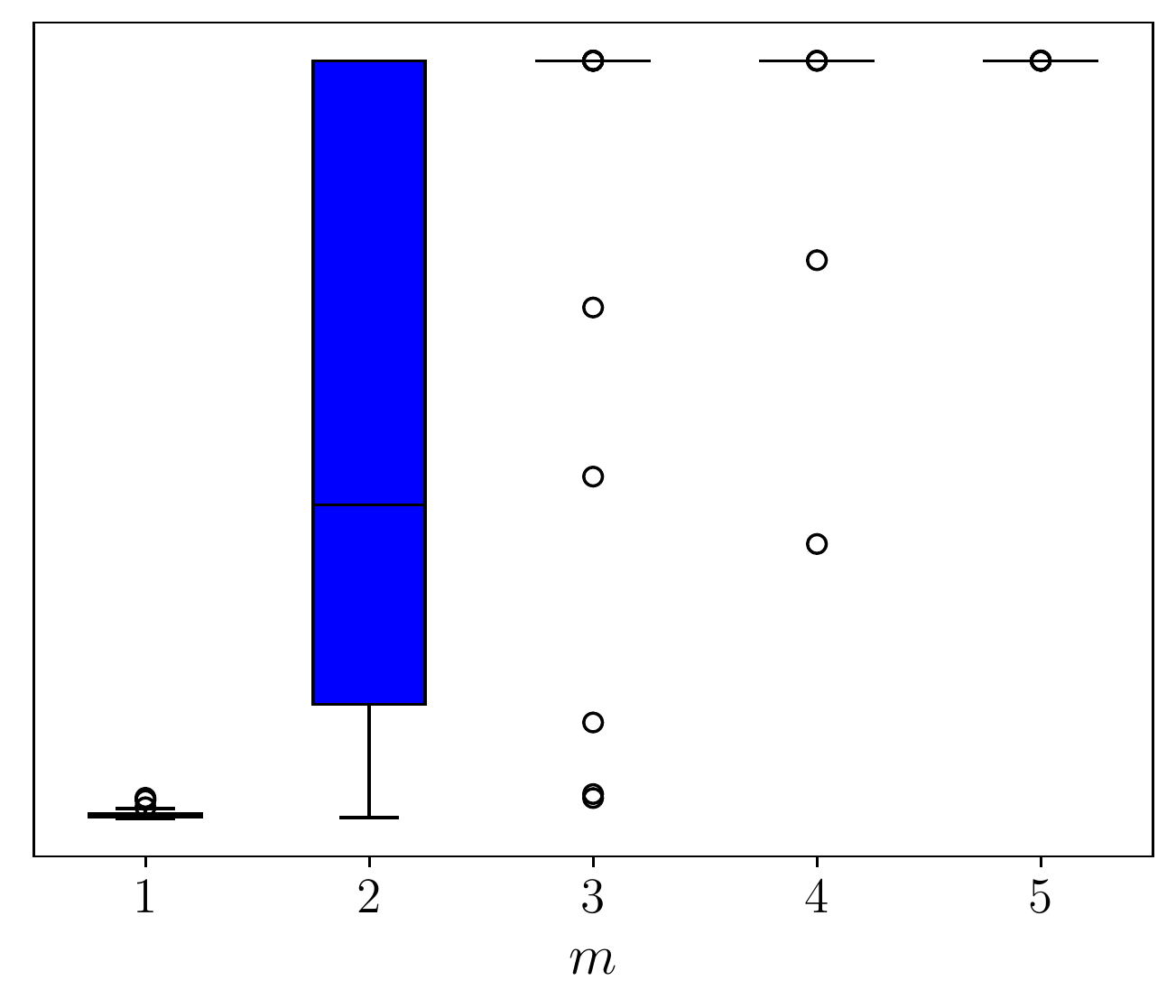}{}
	\caption{Solution times with centralized approach, $h=10$ and varying $m$. Left: OFFICE. Center: GRID-NOFN. Right: GRID-FN.}
	\label{fig:m_times}
\end{figure}

\begin{figure}[ht]
	\centering
	\includegraphics[width=\mywa, height=\myha]{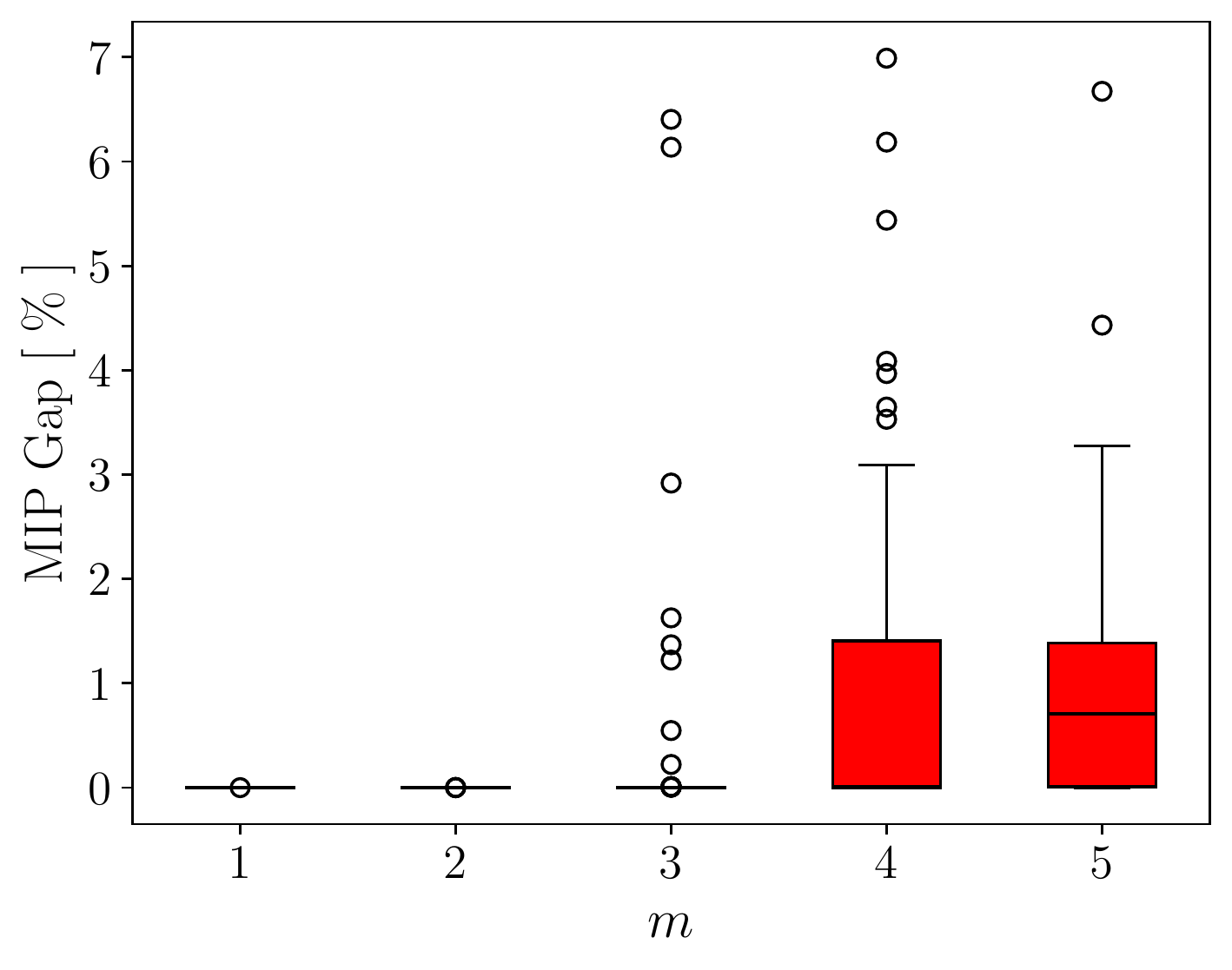}{}
	\includegraphics[width=\mywb, height=\myha]{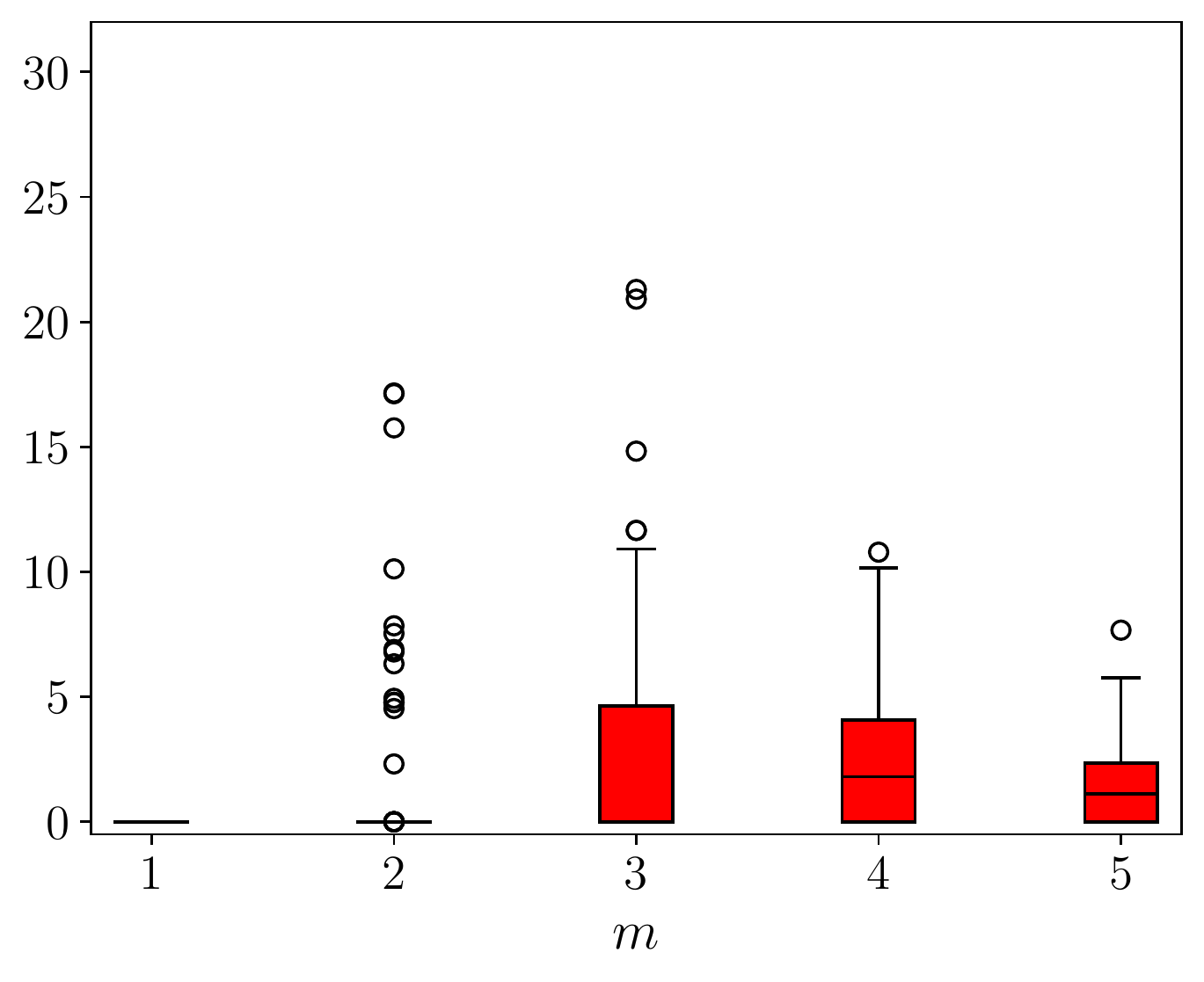}{}
	\includegraphics[width=\mywc, height=\myha]{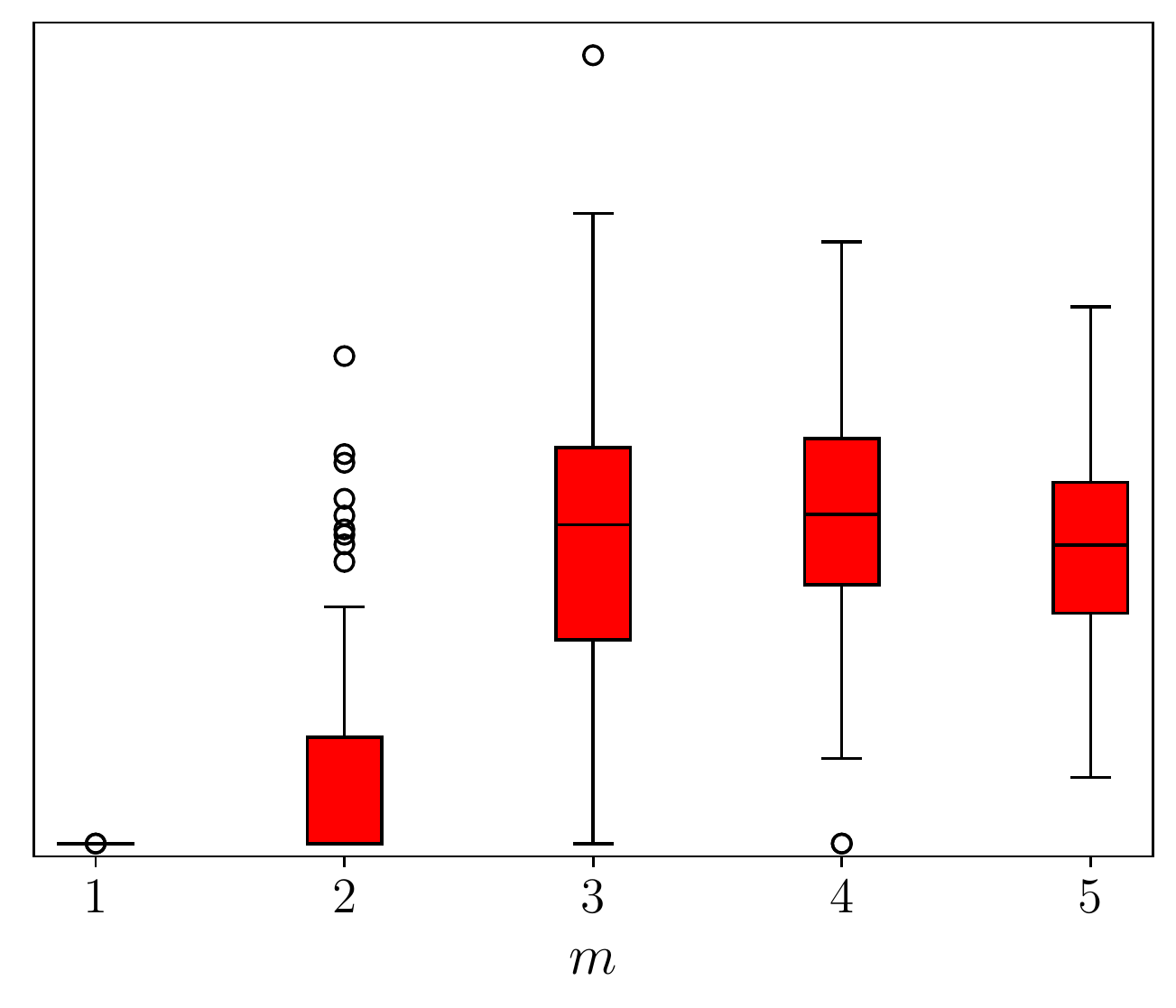}{}
	\caption{MIP gaps with centralized approach, $h=10$ and varying $m$. Left: OFFICE. Center: GRID-NOFN. Right: GRID-FN.}
	\label{fig:m_gaps}
\end{figure}

We choose a planning horizon of $h=10$, shown previously to be optimally solvable within our time limit for a single searcher in all instances. Figs.~\ref{fig:m_times}-\ref{fig:m_gaps} show the solution times and corresponding MIP gaps for OFFICE, GRID-NOFN and GRID-FN. 


A larger search team increases the model complexity (see Fig. \ref{fig:m_times}), and consequently even OFFICE instances hit the time limit for $m\geq3$. These sub-optimal solutions, however, present small MIP gap values ($\leq 7\%$), with median gaps of less than $0.8\%$ (Fig. \ref{fig:m_gaps}, left). On GRID-NOFN, median gap values are still low ($<1.2\%$), although we have some outliers (max. $22\%$). Overall higher gaps are found on GRID-FN, with median values around $12\%$ for $m\geq3$ (Fig. \ref{fig:m_gaps}, right). As it becomes comparatively easier to intercept the \target~with an extra searcher in an environment of this size, the higher MIP gaps in both GRIDs arise when $m=3, 4$. 

\begin{figure}[h!]
	\centering
	\includegraphics[width=\myws, height=\myhb]{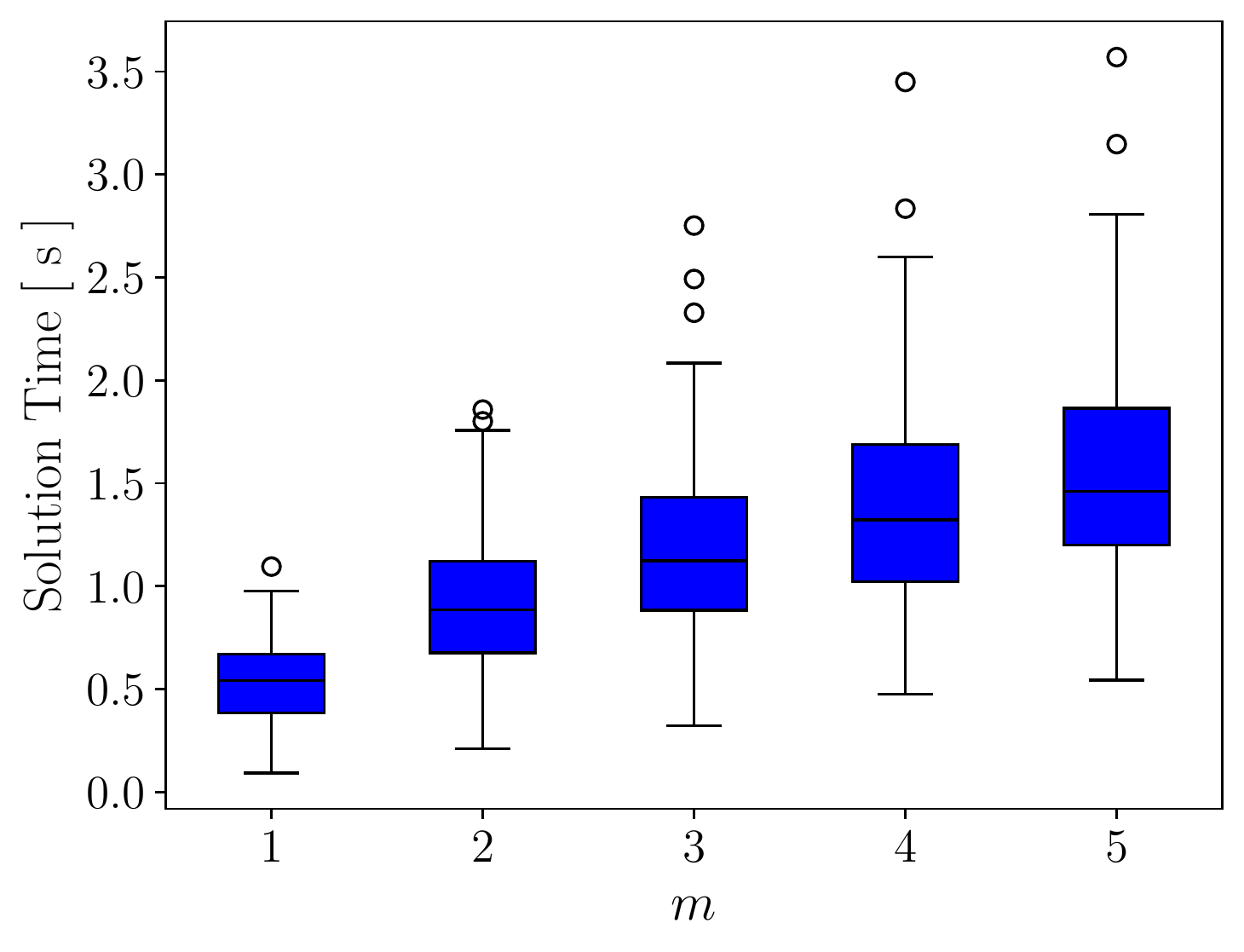}{}
	\includegraphics[width=\myws, height=\myhb]{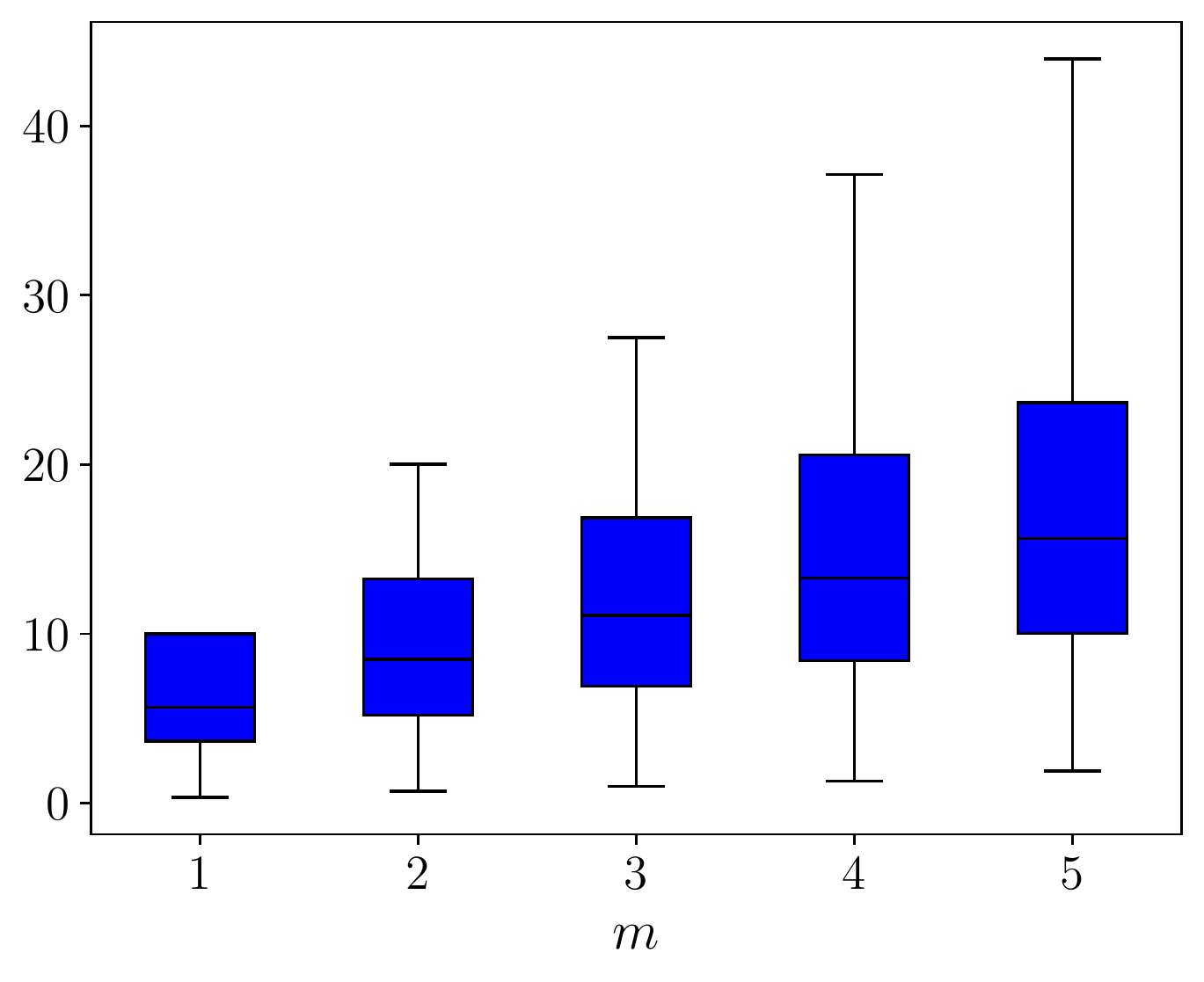}{}
	\caption{Solution times with distributed approach, $h=10$ and varying $m$. Left: OFFICE. Right: GRID-FN.}
	\label{fig:distributed_time}
\end{figure}

% ----------------------------------------------------
% SET 3 distributed, m ={1,..5}, h=10
% ----------------------------------------------------
\textit{ 3a) Scalability of the MILP models for distributed approach w.r.t. team size:} We implement the implicit coordination algorithm described in  Sec. \ref{sec:distributed}, replanning at each time step. Figure \ref{fig:distributed_time} shows the solution times with a planning horizon $h=10$ for OFFICE and GRID-FN.

The distributed solution presents a significantly better scalability than the centralized approach under equivalent conditions, which can be seen by comparing the solution times in Fig. \ref{fig:distributed_time} (left, right) and Fig.~\ref{fig:m_times} (respectively left, right). The following results from OFFICE illustrate this claim: although for $m=1$ the median computational times of the centralized and distributed approaches are similar, respectively $0.52$ sec and $0.54$ sec, for $m=5$ these values increase to approximately $1800$ sec (centralized) and $1.46$ sec (distributed). In comparative terms, a 5x increase in the team size caused the median solution time to increase 3x for the distributed algorithm, against a drastic 3400x increase in solution time for the centralized approach.

\begin{figure}[h!]
	\centering
	\includegraphics[width=\myws, height=\myhb]{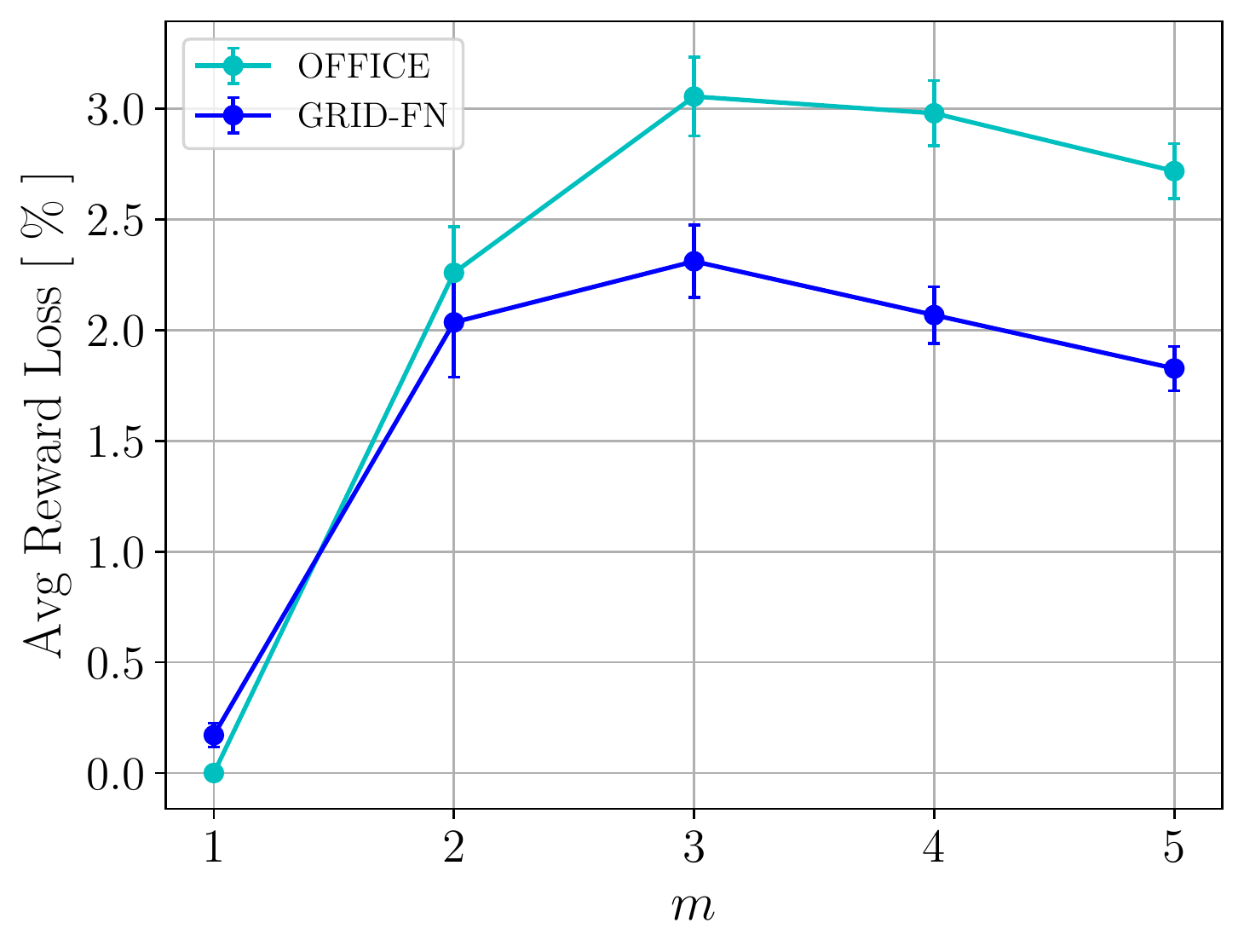}{}
	\includegraphics[width=\myws, height=\myhb]{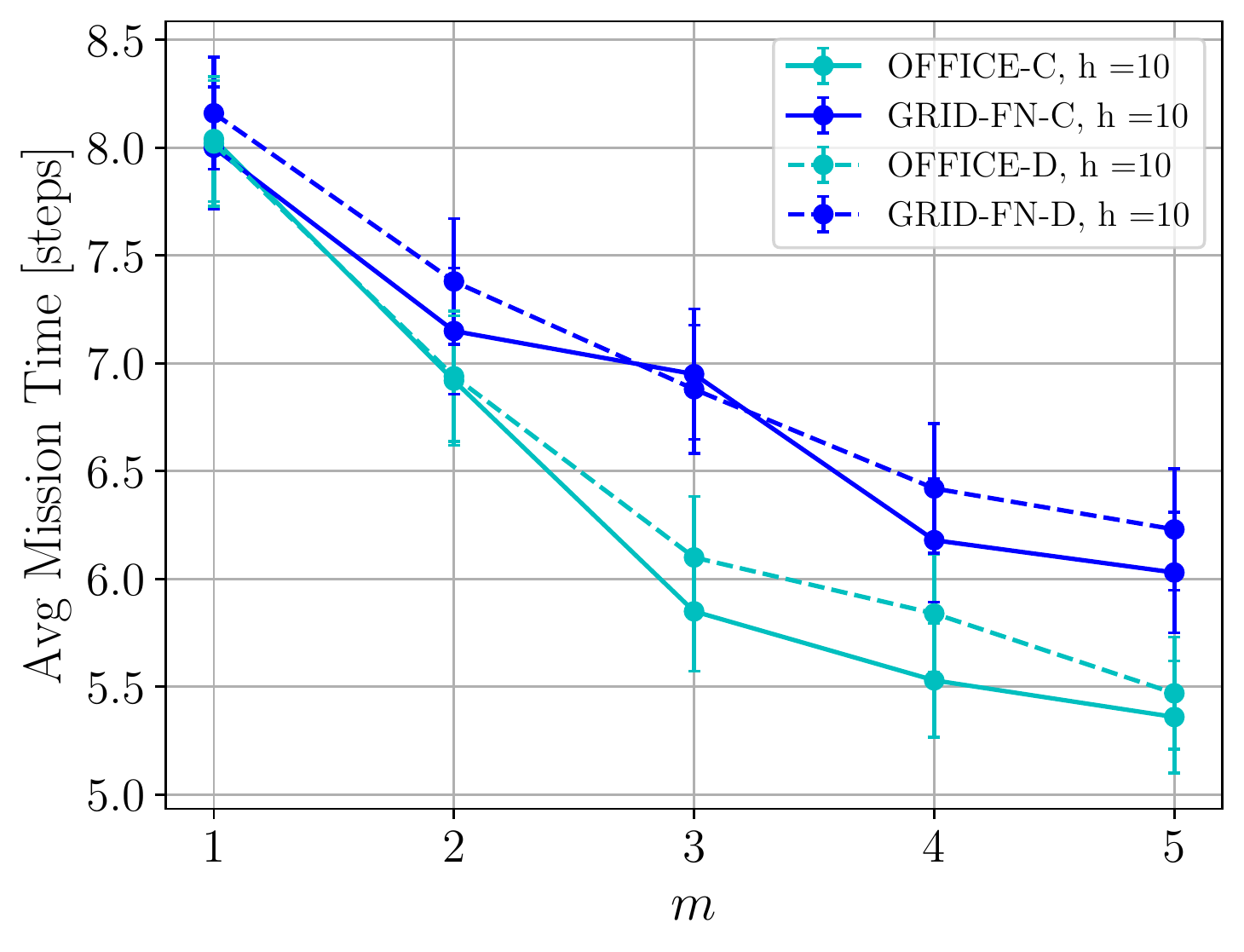}{}
	\caption{Performance of distributed and centralized MILP approaches for OFFICE and GRID-FN, $h = 10$. Left: Relative reward loss computed at $t=0$. Right: Average mission time. Bars show the std. error of the mean.}
	\label{fig:CxDreward}
\end{figure}

\textit{ 3b) Comparative performance of online (distributed) and offline (centralized) MILP search plans:} As basis for comparison, we introduce two metrics: the \textit{average mission time}, defined as the time-step the mission ends due to the expiration of the deadline or capture of the \target; and the \textit{relative reward loss}, defined as the percentage difference between the distributed and centralized reward functions computed at time $t=0$. Figure \ref{fig:CxDreward} shows the relative reward loss (left) and the average mission time (right) for OFFICE and GRID-FN, for a mission deadline $\tau=50$ with $h=10$ and varying $m$.

The relative reward loss for the environments studied in this \paper~is minimal, as shown in Fig. \ref{fig:CxDreward} (left). The higher loss is seen for the OFFICE environment (within 3\% of optimal reward) and slightly lower for GRID-FN (2\% difference). Recall from Fig. \ref{fig:m_times} (right) that the centralized approach often fails to solve the GRID-FN problem to optimality in the time given, which might result in a sub-optimal offline plan, however with a higher reward when compared to the proposed distributed plan. This small difference in reward translates into an overall shorter, if at times irrelevant\footnote{Note for $m=3$ in GRID-FN, the false negative causes the actual detection of the target in the distributed, but not in the centralized instance.}, average mission time for the centralized approach (see Fig. \ref{fig:CxDreward}, right). Given the same planning horizon, the distributed approach often performs nearly as well as the centralized, both w.r.t. reward (within 3\%) and mission time (within 6\%), with the advantage of requiring significant less time.

% ----------------------------------------------------
% SET 4: distributed MILP vs C++, m = 3, h = {2,..8} 
% ---------------------------------------------------
\setcounter{subsubsection}{3}
\subsubsection{Comparison between MILP approach and previous state-of-the-art (SoA) algorithm} 
\begin{figure}[h!]
	\centering
	\includegraphics[width=\myws, height=\myhb]{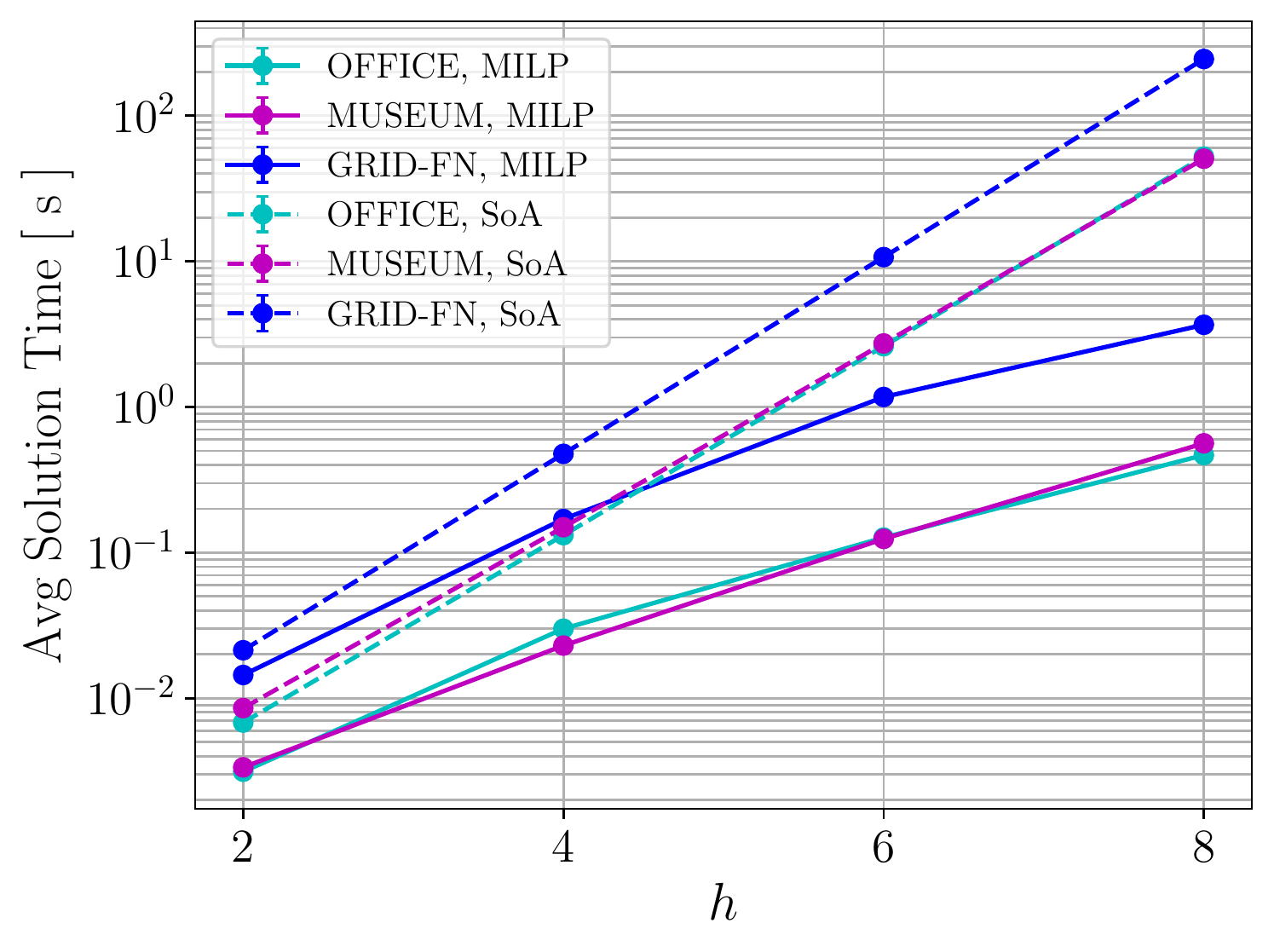}
	\includegraphics[width=\myws, height=\myhb]{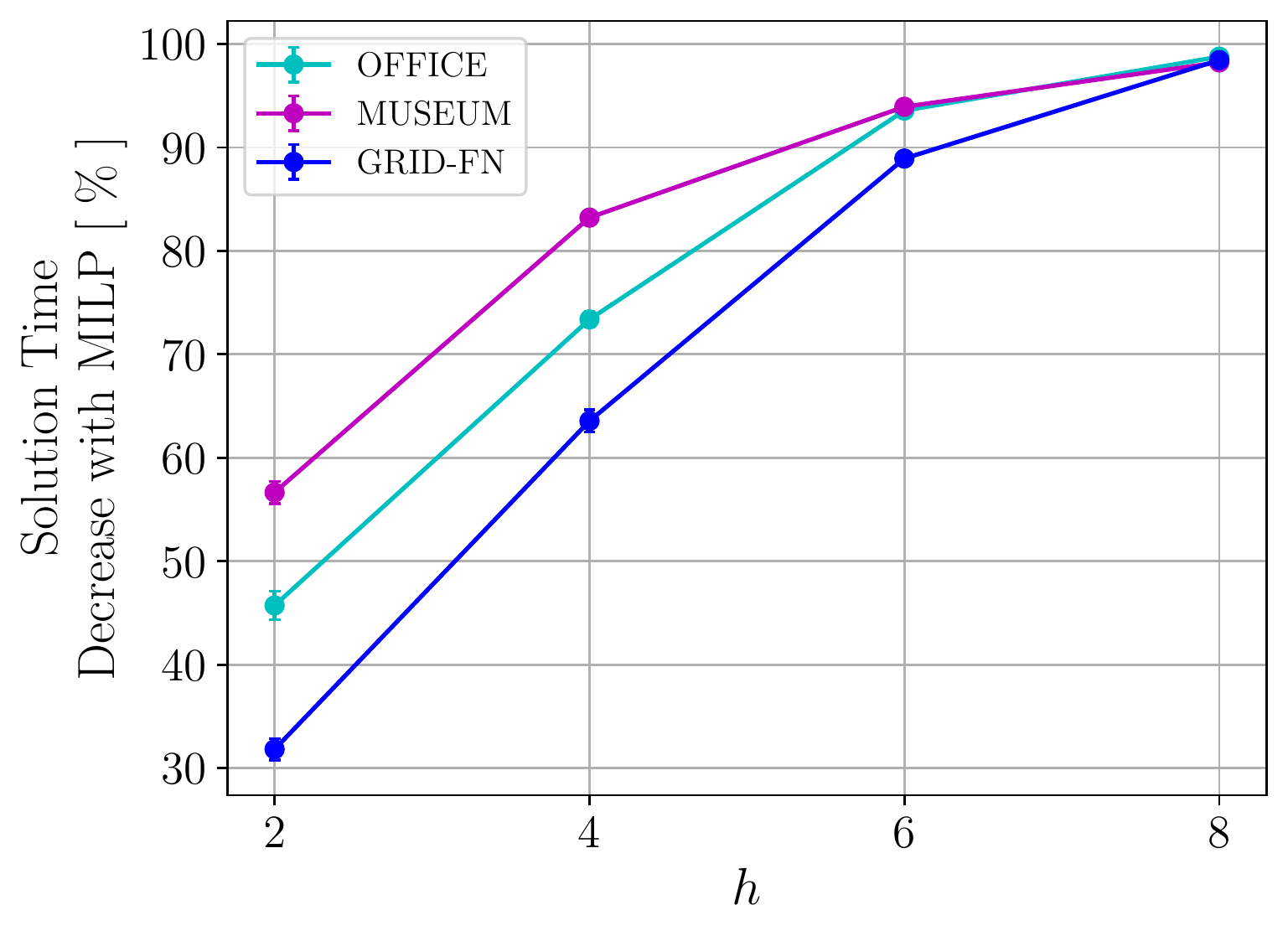}
	\caption{Comparison of MILP and SoA \cite{hollinger2009} distributed approaches for OFFICE, MUSEUM and GRID-FN,  $m=3$ and varying $h$. Left: Average solution time (log scale). Right: Solution time decrease with MILP (relative change with SoA as the reference value). Bars show the std. error of the mean.}
	\label{fig:milpxcpp}
\end{figure}

The solution times for varying planning horizons and $m=3$ are shown in Fig. \ref{fig:milpxcpp} for MUSEUM, OFFICE and GRID-FN.

The implicit coordination algorithm (SoA) was implemented in C++ by the authors\footnote{Code is open source and available at \url{https://github.com/jacoban/implicit\_coordination}.} as presented in~\cite{hollinger2009}. The same machine is used to run  the MILP and SoA experiments, and no time limit is imposed. While the algorithms provide interchangeable solutions (same computed reward), the computational time required to do so varies greatly between them. The MILP paradigm outperforms the previous SoA w.r.t. computational time in all cases, and this difference becomes more expressive as the planning horizon increases  (see Fig. \ref{fig:milpxcpp}, left). {In average terms, for $h=8$ and $m=3$ in the environments tested in this \paper, the MILP models allow for a solution time decrease of 98\% compared to the previous SoA (see Fig. \ref{fig:milpxcpp}, right)}.

% ----------------------------------------------------
% SET 5: distributed MILP m = {1,..,5}, h = {5, 10}
% ----------------------------------------------------
\subsubsection{Performance of the MILP distributed approach with different planning horizons} 
\begin{figure}[h!]
	\centering
	\includegraphics[width=\myws, height=\myhb]{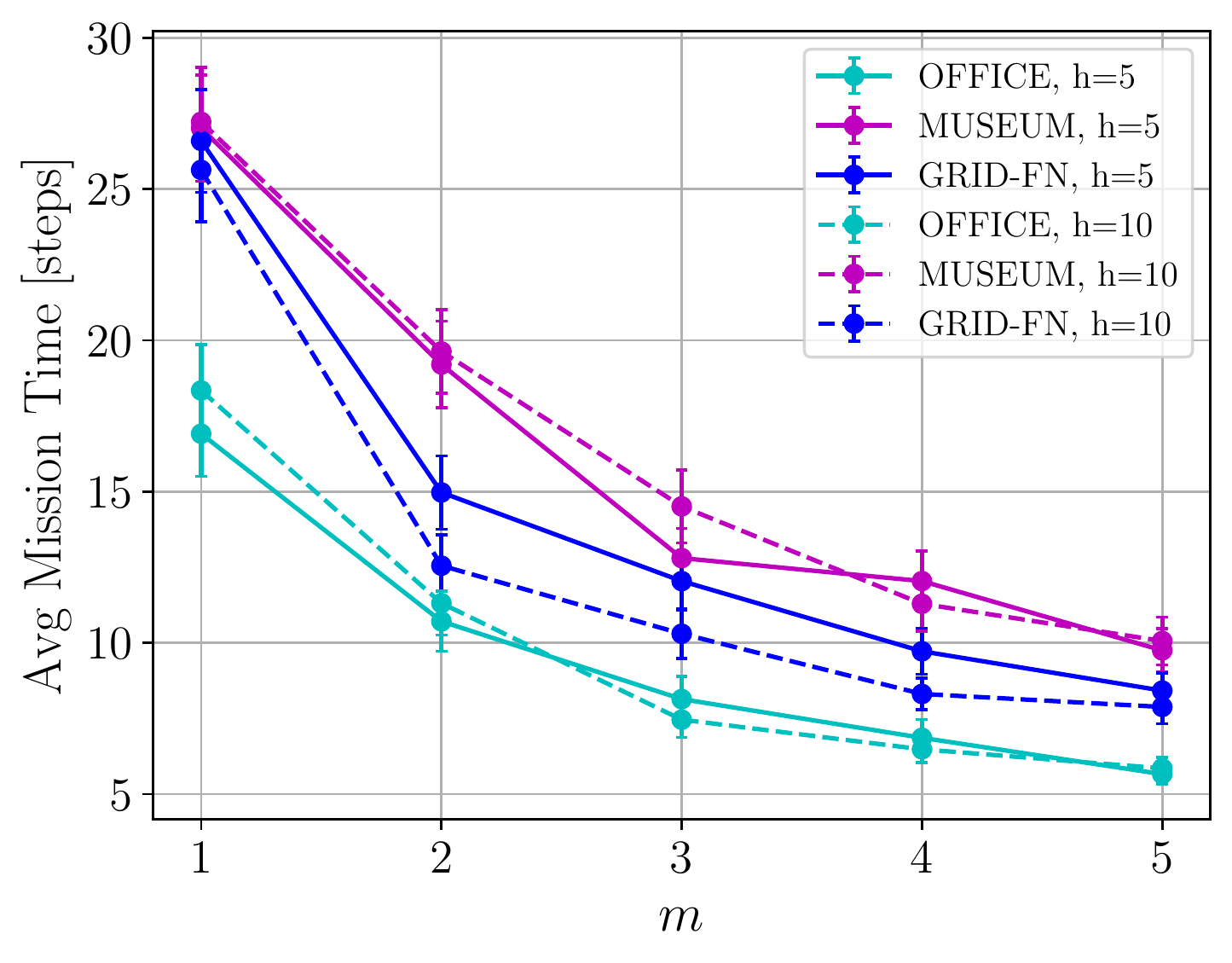}{}
	\includegraphics[width=\myws, height=\myhb]{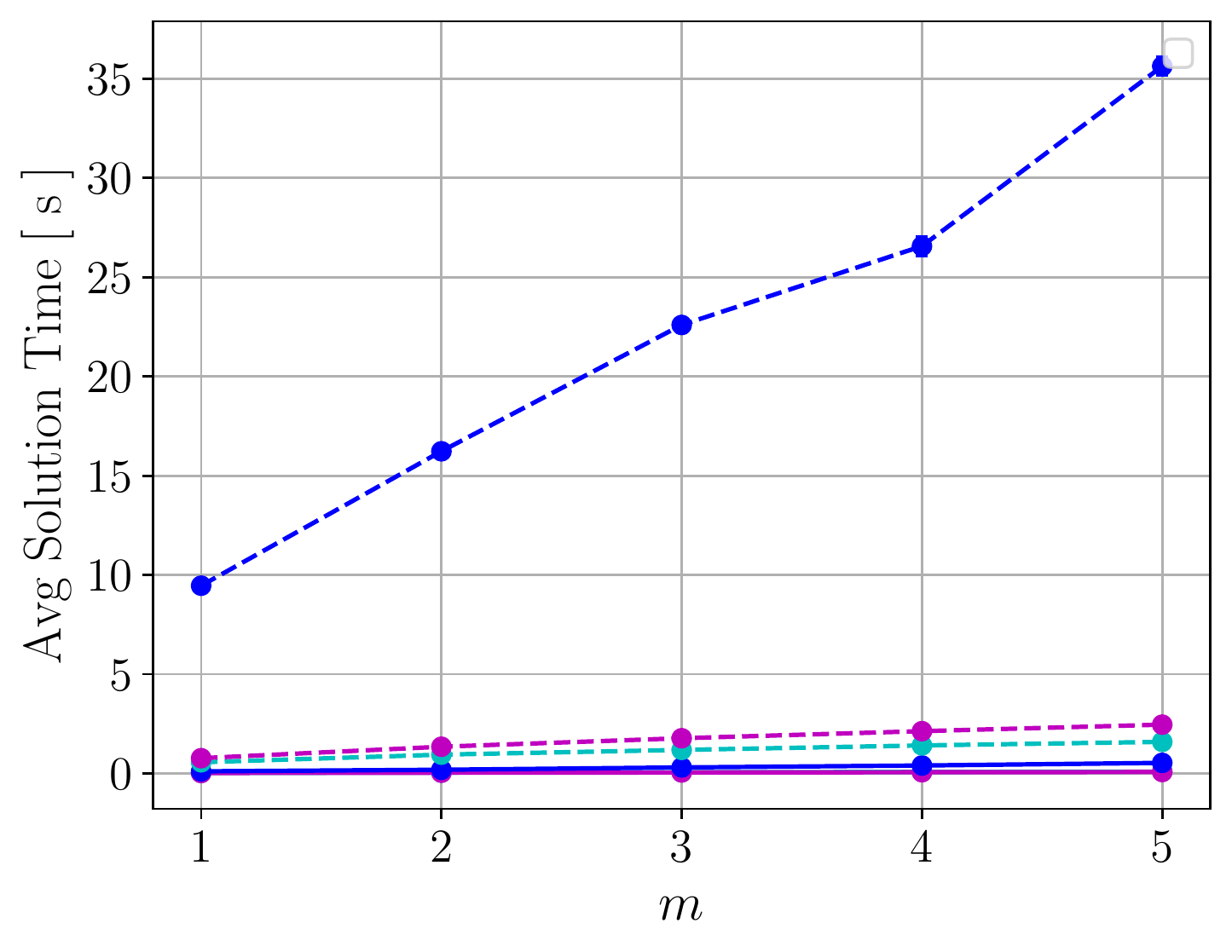}{}
	\caption{Performance of MILP distributed approach, $h = 5, 10$ for OFFICE, MUSEUM and GRID-FN. Left: Average mission time. Right: Average solution time. Bars show the standard error of the mean.}
	\label{fig:DH5xDH10}
\end{figure}

Figure \ref{fig:DH5xDH10} shows the average mission time for $\tau=50$ and the solution time with $h =5, 10$ for OFFICE, MUSEUM and GRID-FN. 

For both OFFICE and MUSEUM there is virtually no difference in performance for the planning horizons tested (Fig. \ref{fig:DH5xDH10}, left). For GRID-FN, the imposed restriction on searchers and \target's initial positions (Sec. \ref{sec:setupsim}) creates a more challenging planning scenario given their relative initial distance. For this case, a longer planning horizon yields better performance, however at the expense of a greater computing time, which grows expressively with the number of searchers (see Fig. \ref{fig:DH5xDH10}, right).

\section{Discussion}
\label{sec:conclusion}

In this \paper, we proved the MESPP problem to be NP-hard even on seemingly simple instances, i.e. grid graphs, static target, and single searcher. We also presented the first set of MILP models able to encompass multiple searchers, arbitrary capture ranges, and false negatives simultaneously. Our results show that the adoption of MILP as a planning paradigm outperforms the previous state-of-the-art approach, both in terms of planning horizon and computational performance.

Leveraging the powerful techniques and tools used by modern solvers comes with a minor challenge: very rarely (three instances in total), the presolver might deal poorly with small probabilities and deem the problem infeasible. This numerical issue is fixed either by turning the presolver off and increasing the solver timeout, or by keeping the searchers' in their current positions and re-planning on next time-step (always a feasible solution and the one we adopted). We believe this is just a small inconvenience, given the benefits provided by the MILP models.

As shown in our simulations, the trade-off between expected mission time and required computational time is a challenging choice. Specially for practical situations, such choice is dependent upon the desired search mission's goals and is fundamental for its success. Future work will continue investigating MILP as a planning paradigm, towards the generalization of the presented models to handle heterogeneous teams of searchers (humans, ground and aerial vehicles), and the incorporation of connectivity constraints.

\section*{Acknowledgment}
Funding for this research was provided by the National Robotics Initiative Program of the National Science Foundation, award number \#1830497 and PERISCOPE MURI Grant N00014-17-1-2699.

\bibliographystyle{IEEEtran}
\bibliography{IEEEabrv,mybib}

\end{document}